\documentclass[a4paper,fleqn]{cas-dc}

\usepackage[numbers, square, comma]{natbib}
\usepackage{float}
\usepackage[framemethod=TikZ]{mdframed}
\usepackage{xcolor}
\usepackage{algorithm}
\usepackage{algpseudocode}
\usepackage{diagbox}
\usepackage{multirow}
\usepackage{booktabs}
\usepackage{amsmath}
\def\tsc#1{\csdef{#1}{\textsc{\lowercase{#1}}\xspace}}
\tsc{WGM}
\tsc{QE}
\tsc{EP}
\tsc{PMS}
\tsc{BEC}
\tsc{DE}

\begin{document}
\let\WriteBookmarks\relax
\def\floatpagepagefraction{1}
\def\textpagefraction{.001}
\let\printorcid\relax

\title [mode = title]{MetaOcc: Spatio-Temporal Fusion of Surround-View 4D Radar and Camera for 3D Occupancy Prediction with Dual Training Strategies}

\author[1]{Long Yang}\fnref{co-first}
\ead{loongyang567@163.com}
\author[1]{Lianqing Zheng}\fnref{co-first}
\ead{zhenglianqing@tongji.edu.cn}
\author[1]{Wenjin Ai}
\author[2]{Minghao Liu}   
\author[3]{Sen Li}
\author[4]{Qunshu Lin}
\author[5]{Shengyu Yan}
\author[6]{Jie Bai}
\author[1]{Zhixiong Ma}\corref{cor1}
\ead{mzx1978@tongji.edu.cn}
\author[7]{Tao Huang}
\author[1]{Xichan Zhu}

\affiliation[1]{organization={School of Automotive Studies, Tongji University},
    city={Shanghai},
    postcode={201804},
    country={China}}                
\affiliation[2]{organization={2077AI Foundation},
    country={Singapore}} 
\affiliation[3]{organization={Autonomous Driving Development, NIO},
    city={Shanghai},
    postcode={201804},
    country={China}} 
\affiliation[4]{organization={College of Computer Science and Technology, Zhejiang University},
    city={Hangzhou},
    postcode={310027},
    country={China}} 
\affiliation[5]{organization={School of Automobile,  Chang'an University},
    city={Xi'an},
    postcode={710018},
    country={China}} 
\affiliation[6]{organization={School of Information and Electrical Engineering, Hangzhou City University},
    city={Hangzhou},
    postcode={310015},
    country={China}} 
\affiliation[7]{organization={College of Science and Engineering, James Cook University},
    city={Cairns},
    country={Australia}}

\fntext[co-first]{Equal Contribution}
\cortext[cor1]{Corresponding author}
\nonumnote{
}

\begin{abstract}
Robust 3D occupancy prediction is essential for autonomous driving, particularly under adverse weather conditions where traditional vision-only systems struggle. While the fusion of surround-view 4D radar and cameras offers a promising low-cost solution, effectively extracting and integrating features from these heterogeneous sensors remains challenging. This paper introduces MetaOcc, a novel multi-modal framework for omnidirectional 3D occupancy prediction that leverages both multi-view 4D radar and images. To address the limitations of directly applying LiDAR-oriented encoders to sparse radar data, we propose a Radar Height Self-Attention module that enhances vertical spatial reasoning and feature extraction. Additionally, a Hierarchical Multi-scale Multi-modal Fusion strategy is developed to perform adaptive local-global fusion across modalities and time, mitigating spatio-temporal misalignments and enriching fused feature representations. 
To reduce reliance on expensive point cloud annotations, we further propose a pseudo-label generation pipeline based on an open-set segmentor. This enables a semi-supervised strategy that achieves 90\% of the fully supervised performance using only 50\% of the ground truth labels, offering an effective trade-off between annotation cost and accuracy.
Extensive experiments demonstrate that MetaOcc under full supervision achieves state-of-the-art performance, outperforming previous methods by +0.47 SC IoU and +4.02 mIoU on the OmniHD-Scenes dataset, and by +1.16 SC IoU and +1.24 mIoU on the SurroundOcc-nuScenes dataset.
These results demonstrate the scalability and robustness of MetaOcc across sensor domains and training conditions, paving the way for practical deployment in real-world autonomous systems. Code and data are available at \url{https://github.com/LucasYang567/MetaOcc}.
\end{abstract}


\maketitle

\section{Introduction}
Environmental perception plays a pivotal role in advancing autonomous driving systems \cite{ma2024cam4docc, SHI2025113539}.
Compared to 3D object detection, 3D occupancy prediction enables the modeling of arbitrarily shaped objects, thereby supporting finer-grained scene understanding and more accurate interpretation of complex environments \cite{ding2024radarocc}.
Although surround-view cameras have been widely adopted in recent research for their affordability and rich semantic information, they exhibit limited depth perception and reduced robustness under adverse weather conditions \cite{xu2025occsurvey, xiong2023lxl, yuan2025if}.

To address these limitations, the fusion of surround-view cameras with cost-effective 4D millimeter-wave radar has emerged as a promising approach for robust and accurate occupancy prediction \cite{OmniHDScenes}.
In addition to the traditional benefits of radar, such as resilience to environmental factors and low cost, 4D radar introduces height information and produces denser point clouds, enabling improved spatial reasoning.
The fusion of 4D radar with surround-view cameras thus facilitates complementary sensing, offering omnidirectional perception with both semantic and geometric precision \cite{smurf, WaterVG, zheng2023rcfusion}.
However, research on 4D radar-camera fusion specifically for occupancy prediction remains limited.

Recent studies typically perform fusion at the representation level by projecting features from both modalities into a Bird’s Eye View (BEV) or voxel space \cite{co-occ, wang2023openoccupancy, zhang2023occformer, xu2024mrftrans}.
These methods often rely on simple operations such as concatenation or summation to merge features \cite{zheng2023rcfusion, Talk2Radar}.
Such approaches lack the flexibility to adaptively weight modality contributions and often fail to resolve spatio-temporal misalignment caused by asynchronous sensor sampling.
Moreover, many fusion methods repurpose LiDAR-oriented encoders (e.g., PointNet \cite{RN175}, PointPillars \cite{Pointpillars}, or VoxelNet \cite{RN177}) for radar processing.
Given the fundamental differences in sensing principles and the inherent sparsity of radar point clouds, this direct transfer is often suboptimal for radar-specific feature extraction.

In addition, most current methods rely on densely annotated ground-truth occupancy labels, which require labor-intensive point-level segmentation \cite{OmniHDScenes, wang2023openoccupancy, surroundocc}.
Recent self-supervised approaches aim to reduce annotation demands by leveraging neural rendering or geometric consistency \cite{zhang2023occnerf, liu2024let}.
Although these methods mitigate reliance on labeled data, they are often computationally intensive and struggle to accurately capture dynamic elements in complex scenes, limiting their practical deployment.

\begin{figure}[t]
    \centering
    \includegraphics[width=\linewidth]{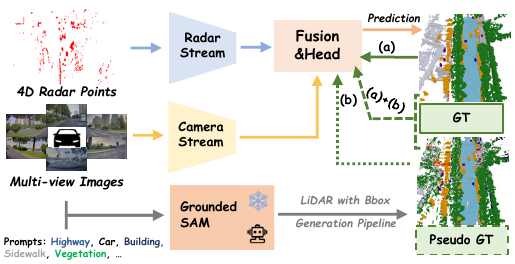}
    \caption{
    Overview of the proposed occupancy prediction framework leveraging surround-view 4D radar and camera fusion for robust 3D scene understanding. (a) Fully supervised training using ground-truth labels. (b) Weakly supervised training with only pseudo-labels. (a + b) Semi-supervised training that integrates both ground-truth and pseudo-labels.}
    \label{fig:overview}
\end{figure}

To address the challenges mentioned above and advance robust 3D scene understanding for autonomous systems, this work introduces MetaOcc, a novel surround-view occupancy prediction framework that integrates 4D radar and camera data through a spatio-temporal fusion architecture. 
By addressing key challenges in radar feature extraction, cross-modal alignment, and annotation efficiency, MetaOcc enables accurate and resilient semantic occupancy prediction in diverse and adverse environments.

Rather than relying on LiDAR-specific encoders, we design a Radar Height Self-Attention module that enhances spatial feature extraction from sparse radar point clouds. We further design a Hierarchical Multi-Scale Multi-Modal Fusion module, which performs efficient spatio-temporal fusion across modalities. To reduce annotation costs, we propose a novel pseudo-label generation pipeline based on zero-shot semantic segmentation models with text prompts.

The main contributions are summarized as follows:
\begin{itemize}
\item 
To the best of our knowledge, MetaOcc is the first framework to fuse surround-view 4D radar and camera data for 3D occupancy prediction. The method generalizes effectively to conventional radar-camera fusion settings and achieves state-of-the-art performance. Specifically, it surpasses previous methods by +0.47 SC IoU and +4.02 mIoU on the OmniHD-Scenes dataset, and by +1.16 SC IoU and +1.24 mIoU on the SurroundOcc-nuScenes dataset.
\item We propose a Radar Height Self-Attention module that effectively extracts fine-grained 3D spatial features from sparse radar point clouds.
By applying self-attention along the height dimension, this module explicitly models vertical structures and improves geometric reasoning in complex scenes, addressing the limitations of applying LiDAR-oriented encoders directly to radar data.
\item We introduce a Hierarchical Multi-Scale Multi-Modal Fusion module, which dynamically adjusts the contribution of radar-camera features, enables global cross-modal interaction via deformable attention, and aggregates temporal information.
Unlike existing approaches that rely on static fusion strategies such as concatenation or summation, the module adaptively integrates multi-frame radar-camera features, addressing spatio-temporal misalignment and improving prediction robustness in complex environments.
\item 
We design a novel pseudo-label generation pipeline that leverages a zero-shot segmentation model guided by text prompts. This pipeline exhibits strong generalizability and effectiveness across different models. 
Notably, models trained solely on pseudo-labels can achieve approximately 65\% of their fully supervised performance. When integrated into a semi-supervised learning strategy using only 50\% of the ground-truth annotations, performance improves to nearly 90\% of the fully supervised level, offering a highly efficient trade-off between annotation cost and accuracy.
\end{itemize}

The remainder of this paper is organized as follows. 
Section~\ref{sec:related_work} reviews related work on 3D occupancy prediction and multi-modal radar-camera fusion techniques. 
Section~\ref{sec:method} presents the MetaOcc framework, including radar and camera feature extraction modules, spatio-temporal fusion mechanisms, and the pseudo-label generation pipeline for semi-supervised learning. 
Section~\ref{sec:experiments} describes the experimental setup and reports quantitative and qualitative results, including robustness evaluations and ablation studies.
Finally, Section~\ref{sec:conclusions} concludes the paper and outlines potential directions for future research.

\section{Related Work}
\label{sec:related_work}
\subsection{Camera-based 3D Occupany Prediction}
Surround-view cameras are widely adopted in vehicle perception due to their affordability and the rich semantic information they provide.
In 3D occupancy prediction tasks, most existing methods convert multi-view images into unified feature representations such as Bird’s Eye View (BEV), Tri-Perspective View (TPV), or voxel space \cite{surroundocc, li2023voxformer, li2023fb, FastOcc, TPVFormer, huang2021bevdet, huang2022bevdet4d}.
Among these, BEV provides a flattened spatial representation, which makes it challenging to directly support detailed 3D scene reconstruction.
To address this limitation, several studies have proposed mechanisms to recover 3D spatial features from BEV representations.
FlashOcc \cite{yu2023flashocc} introduces a channel-to-height transformation module that reconstructs voxel features from BEV, enhancing vertical scene understanding.
OccFormer \cite{zhang2023occformer} proposes a dual-branch attention encoder, where local voxel features guide compressed global BEV features to learn spatial weights across heights for improved 3D understanding.

TPVFormer \cite{TPVFormer} extends BEV by integrating two orthogonal vertical planes to form a Tri-Perspective View, significantly boosting prediction performance.
Building on this concept, S2TPVFormer \cite{S2TPVFormer} introduces a hybrid long-term cross-view attention mechanism to capture rich spatio-temporal context across scenes.
Although TPV provides improved geometric representation compared to BEV, it still faces limitations in reconstructing complete 3D scenes from planar features.

Voxel-based approaches, by contrast, directly perform feature learning in 3D space and are generally more suitable for accurate 3D occupancy prediction \cite{tang2024sparseocc}.
Pioneering works such as SurroundOcc \cite{surroundocc} and Occ3D \cite{tian2024occ3d} project multi-view image features into 3D voxels using 2D-to-3D spatial attention and enhance performance through refinement modules.
Recent methods further improve voxel representation by incorporating geometric-semantic awareness \cite{Radocc}, diffusion denoising \cite{wang2025occgen}, and Gaussian splatting techniques \cite{gaussianformer, gaussianformer2}.

Despite the progress made by BEV-, TPV-, and voxel-based camera methods, these approaches remain limited by the inherent drawbacks of camera sensors.
In particular, cameras lack reliable depth information and are highly sensitive to lighting conditions and adverse weather, which significantly affects performance in real-world autonomous driving scenarios.
As a result, camera-only systems often struggle with spatial ambiguity and reduced robustness in low-visibility environments.
To overcome these limitations, we propose incorporating 4D radar features, which offer complementary spatial information such as object height, motion, and robustness to environmental interference.

\subsection{Radar-camera 3D Occupany Prediction}
Millimeter-wave radar has become a key sensor in vehicle perception systems due to its long-range sensing capability, Doppler velocity estimation, and robustness under adverse weather conditions \cite{xiong2023lxl, ming2024occfusion, NanoMVG}.
However, radar point clouds are inherently sparse and suffer from low spatial resolution, making it difficult to generate dense and accurate occupancy predictions when used in isolation.
Consequently, radar is typically employed as an auxiliary modality to support other sensor inputs.

In recent years, the fusion of surround-view millimeter-wave radar and cameras has gained increasing attention, as the complementary strengths of these modalities, geometric robustness from radar and semantic richness from cameras, offer a promising path toward cost-effective and reliable 3D occupancy prediction \cite{ming2024occfusion, LiCROcc, lin2024teocc}.
OccFusion \cite{ming2024occfusion} extracts voxel-level radar features using VoxelNet \cite{RN177} and introduces an attention-based 3D-2D dynamic fusion module to adaptively integrate radar and camera features.
LiCROcc \cite{LiCROcc} leverages a LiDAR-camera fusion network as a teacher model to improve radar-camera fusion through cross-modal distillation.
TEOcc \cite{lin2024teocc} introduces a random dropout mechanism on multi-scale fused features to enhance robustness under uncertain conditions.

Notably, existing radar-camera fusion methods utilize conventional millimeter-wave radar, which lacks elevation (height) information and produces sparser point clouds.
In contrast, 4D radar captures elevation data and generates denser measurements, offering greater potential for 3D scene understanding when fused with camera inputs \cite{smurf, Dualradar}.
Despite these advantages, research exploring occupancy prediction using 4D radar-camera fusion remains very limited \cite{OmniHDScenes}.

Moreover, these approaches often repurpose LiDAR-oriented encoders for radar point clouds without accounting for radar-specific properties such as greater sparsity and noise, which leads to suboptimal feature representations.
Additionally, simple fusion strategies such as concatenation or summation are commonly employed, which lack the adaptability to weight modality-specific reliability and fail to address cross-modal misalignment effectively.

To address this gap, we propose a novel 3D occupancy prediction framework that leverages surround-view 4D radar and camera fusion.
Specifically, a Radar Height Self-attention module is introduced to enhance spatial feature extraction from radar point clouds, and a Local-Global Interaction Attention mechanism is designed to facilitate robust and accurate multi-modal fusion.

\subsection{Supervision Strategies for Occupancy Prediction}
Supervision strategies play a critical role in determining the performance of occupancy prediction models \cite{OmniHDScenes}.
Most existing methods rely on fully supervised learning, using dense ground-truth labels to train models for accurate 3D occupancy estimation \cite{surroundocc, wang2023openoccupancy, tian2024occ3d}.
However, constructing high-quality semantic occupancy labels in 3D space is labor-intensive and technically challenging.
Although pipelines such as those in SurroundOcc \cite{surroundocc}, Occ3D \cite{tian2024occ3d}, and OpenOccupancy \cite{wang2023openoccupancy} automate portions of the labeling process and leverage temporal information to improve label quality, they still rely heavily on semantic segmentation of LiDAR point clouds.

To alleviate the annotation burden, recent research has explored self-supervised approaches \cite{zhang2023occnerf, liu2024let}.
Methods such as GenerOcc \cite{GenerOcc} and GeOcc \cite{GEOcc} reformulate occupancy prediction as a depth estimation task and apply contrastive learning between depth maps across consecutive frames.
While these techniques eliminate the need for manual labels, they often fail to capture semantic features essential for reliable scene understanding.
Other methods, including SelfOcc \cite{Selfocc} and TT-GaussOcc \cite{TT-GaussOcc}, introduce spatio-temporal geometric-semantic consistency via Neural Radiance Fields (NeRF) and Gaussian Splatting (GS) to improve 3D reconstruction and occupancy prediction.
Despite their promise, these self-supervised methods generally incur high computational overhead and face difficulties in accurately predicting dynamic elements in complex environments.

More recently, large-scale text-driven segmentation models have demonstrated strong zero-shot generalization and are being explored for use in autonomous driving tasks \cite{ren2024grounded}.
Inspired by this trend, we adopt such a model to reduce dependence on point cloud-based semantic segmentation for static scenes.
By integrating a vision-language segmentation model with 3D bounding box (Bbox) annotations, we generate pseudo-occupancy labels.
On this basis, we develop a cost-effective semi-supervised training strategy that progressively incorporates a small proportion of human-annotated ground truth to guide model optimization.
In comparison to fully supervised and self-supervised approaches, our method strikes a practical balance between annotation efficiency and computational scalability, offering an effective and adaptable solution for large-scale occupancy prediction, as illustrated in Fig.~\ref{fig:overview}.

\begin{figure*}[ht]
    \centering
    \includegraphics[width=\linewidth]{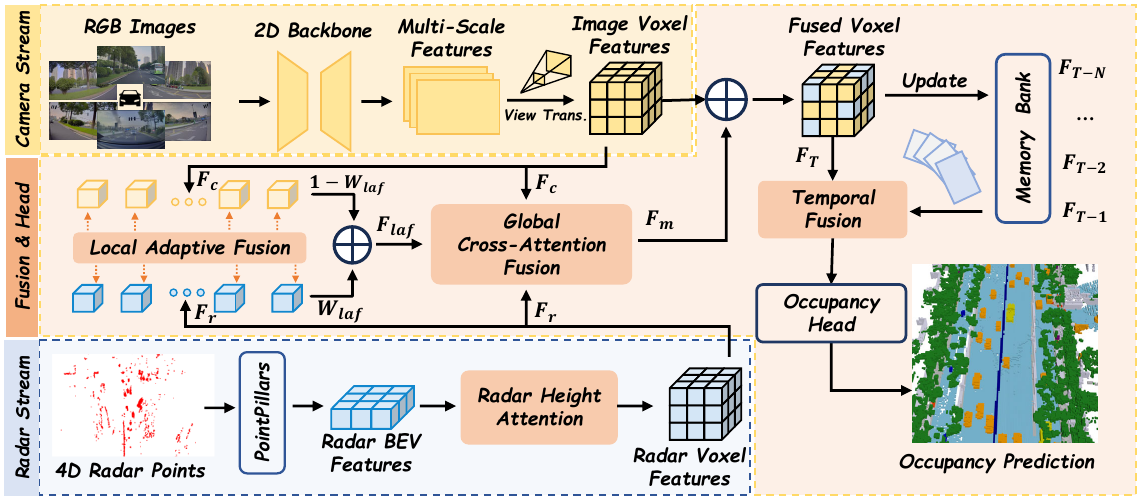}
    \caption{
    Architecture of the proposed MetaOcc framework. (a) Multi-view images are encoded into 3D features via an image feature extractor. (b) Multi-view 4D radar point clouds are voxelized using PointPillars and further enhanced by the Radar Height Self-Attention module. (c) Local Adaptive Fusion and Global Cross-Attention Fusion modules jointly perform efficient fusion of spatial features across modalities. (d) The Temporal Fusion module temporally aligns and aggregates multi-modal features from multiple frames to produce the final occupancy predictions.
    }
\label{fig:MetaOcc Framework}
\end{figure*}

\section{Method}
\label{sec:method}
\subsection{Overall Architecture}
In this work, we propose MetaOcc, a surround-view 3D occupancy prediction framework that fuses 4D millimeter-wave radar and camera inputs.
The overall architecture of the proposed framework is illustrated in Fig.~\ref{fig:MetaOcc Framework}.
Multi-view images and radar point clouds are first processed by modality-specific encoders to extract spatial voxel-level features.
These multi-modal features are subsequently fed into a Hierarchical Multi-Scale Multi-Modal Fusion module, which performs efficient spatio-temporal fusion across modalities to generate the final occupancy prediction.

To mitigate the cost associated with manual annotation of static elements in 3D occupancy ground-truth labels, we incorporate a text-driven open-set segmentator to automatically generate pseudo labels.
These pseudo labels are used as supervision signals during training to reduce reliance on point-level annotated LiDAR data.
The pipeline for pseudo-label generation is depicted in Fig.~\ref{fig:SemiOcc_Pipeline}.

\subsection{Camera Stream}
\textbf{Image Encoder.}
Given multi-view images as input, we adopt ResNet \cite{he2016resnet} as the 2D backbone, integrated with a Feature Pyramid Network (FPN) \cite{lin2017fpn} to extract multi-scale image features.
The resulting feature representation is denoted as $F_I \in \mathbb{R}^{N_c \times C_I \times H_I \times W_I}$, where $N_c$ is the number of cameras, and $C_I$, $H_I$, and $W_I$ refer to the channel, height, and width of the image feature maps, respectively.

\textbf{2D-to-3D View Transformation.}
Efficiently converting 2D image features into 3D spatial voxel features is the core functionality of the Camera Stream.
Traditional forward projection methods based on depth estimation \cite{lss, huang2021bevdet} often struggle with occlusions and depend heavily on ground-truth depth supervision.
To address these limitations, recent attention-based view transformation approaches \cite{bevformer, surroundocc} initialize 3D reference points in the voxel space and project them onto the image plane.
A deformable attention mechanism is then applied to dynamically aggregate 2D image features, enabling more flexible and accurate reconstruction of spatial features.

Building on this, we apply spatial attention-based view transformation at multiple FPN feature scales.
To further enhance spatial resolution, we employ a module composed of 3D convolution and deconvolution layers for voxel feature upsampling and fusion.
This process yields fine-grained voxelized image features denoted as $F_c \in \mathbb{R}^{C \times H \times W \times Z}$, where $W$, $H$, and $Z$ correspond to the dimensions of the occupancy ground truth.

\begin{figure}[t]
    \centering
    \includegraphics[width=\linewidth]{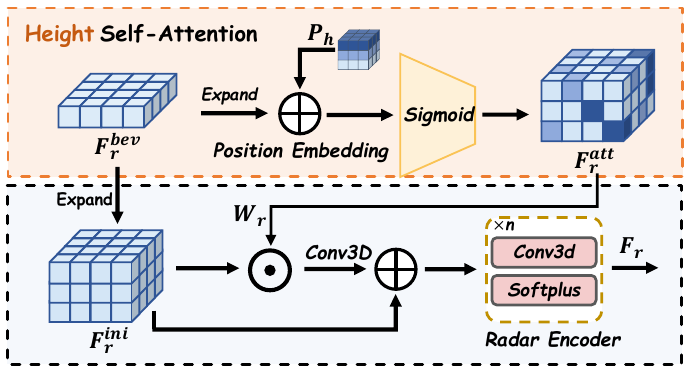}
    \caption{
    Architecture of the proposed Radar Height Self-Attention module, designed to enhance spatial feature extraction from sparse 4D radar point clouds by explicitly modeling vertical structure and height-aware interactions.
    }
    \label{fig:RHS}
\end{figure}

\subsection{4D Radar Stream}

Most existing camera-radar fusion methods employ voxel encoders \cite{yan2018second, RN177} originally designed for LiDAR to process radar point clouds.
However, due to the increased sparsity and noise in radar data, these voxel-based encoders are suboptimal for 3D occupancy prediction tasks.
In contrast, pillar-based encoders \cite{Pointpillars, PillarNeSt} demonstrate stronger performance under sparse conditions but often struggle to model vertical structures effectively.
To address this limitation, we introduce a Radar Height Self-Attention module that enhances voxel feature extraction from radar data using a pillar-based backbone.

\textbf{Radar Pillar Encoder.}
We adopt the classic PointPillars method \cite{Pointpillars} to extract Bird’s Eye View (BEV) features from radar point clouds.
Let $P$ denote the number of non-empty pillars, and $N$ the number of radar points sampled within each pillar.
A multi-layer perceptron is applied to encode point-level features, followed by a max-pooling operation that aggregates features within each pillar.
This process yields BEV radar features denoted as $F_p^{bev} \in \mathbb{R}^{C \times H \times W}$.

\textbf{Radar Height Self-Attention (RHS).}
While feature expansion operations and channel-to-height (C2H) mappings can recover 3D spatial resolution from dense BEV features, they are unable to model vertical feature distributions effectively.
To overcome this, we propose the Radar Height Self-Attention module, which augments BEV features with height-aware contextual information to support effective 3D occupancy reasoning, as shown in Fig.~\ref{fig:RHS}.

First, the BEV feature $F_p^{bev}$ is expanded along the Z-axis to initialize voxel features:
\begin{equation}
F_r^{ini} = \mathtt{Expand}(F_p^{bev}, Z) \in \mathbb{R}^{C \times H \times W \times Z}
\end{equation}
where $\mathtt{Expand}(\cdot, Z)$ denotes repeating the input BEV features $Z$ times along the vertical axis.

To inject vertical awareness, we introduce a learnable height positional encoding $P_h \in \mathbb{R}^{C \times Z}$, which is broadcast-expanded to match the full 3D feature shape, resulting in $P_h^e \in \mathbb{R}^{C \times H \times W \times Z}$.
By summing $P_h^e$ with $F_r^{ini}$, the network gains sensitivity to vertical context, enabling better distinction among objects with different heights (e.g., vehicles, pedestrians, and vegetation) under a unified BEV signature.

The resulting feature is processed by a multi-layer 3D convolutional network $\mathcal{G}_r$, followed by a sigmoid activation $\sigma(\cdot)$ to produce adaptive attention weights.
These weights are used to modulate the original voxel features $F_r^{ini}$, and a final 3D convolutional layer is applied:
\begin{equation}
F_r^{att} = \mathtt{Conv}\left(F_r^{ini} \odot \sigma\left(\mathcal{G}_r\left(F_r^{ini} + P_h^e\right)\right)\right)
\end{equation}
where $\odot$ denotes element-wise multiplication, and $\mathtt{Conv}(\cdot)$ is a 3D convolution operator.

Finally, we apply a residual connection between the initial and attention-enhanced features, followed by a RadarEncoder module composed of stacked 3D convolutional blocks with Softplus activations, yielding the final radar voxel features $F_r$:
\begin{equation}
F_r = \mathtt{RadarEncoder}(F_r^{ini}+F_r^{att}).
\end{equation}

\begin{figure}[t]
    \centering
    \includegraphics[width=0.95\linewidth]{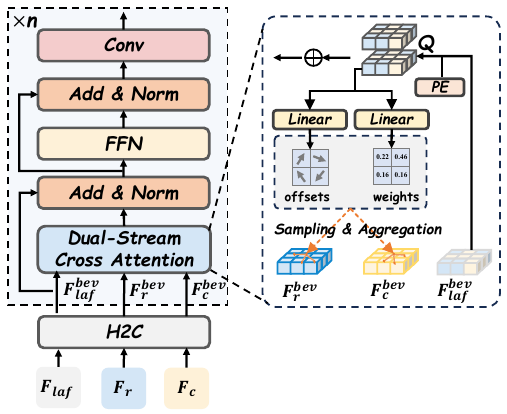}
    \caption{
    Architecture of the Global Cross-Attention Fusion module. Multi-modal 3D features are projected to BEV using a height-to-channel (H2C) operator and fused via deformable cross-attention to address spatial and temporal misalignments across modalities.
    }
    \label{fig:gcf}
\end{figure}

\begin{figure}[t]
    \centering
    \includegraphics[width=0.9\linewidth]{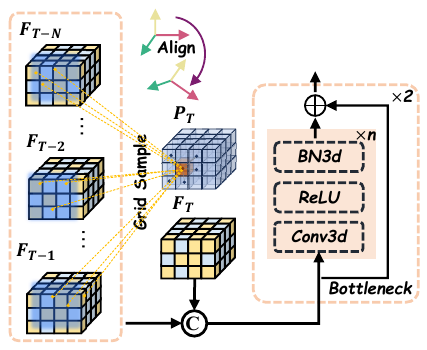}
    \caption{
    Architecture of the Temporal Fusion module. The module efficiently aligns and integrates 3D multi-modal features across consecutive time frames, enhancing temporal consistency for robust occupancy prediction.
    }
    \label{fig:TAF}
\end{figure}

\subsection{Hierarchical Multi-scale Multi-modal Fusion} 

The proposed Hierarchical Multi-scale Multi-modal Fusion module integrates multi-modal information across space, time, and scale to support robust 3D occupancy prediction.
It comprises three core components: Local Adaptive Fusion, Global Cross-Attention Fusion, and Temporal Fusion.
These modules are designed to complement one another by adaptively merging local spatial details, capturing long-range cross-modal dependencies, and incorporating temporal context from sequential frames.
Together, they enable comprehensive and flexible multi-modal spatio-temporal fusion for improved occupancy estimation.

\textbf{Local Adaptive Fusion (LAF).}
Multi-modal features provide complementary strengths across diverse driving scenarios.
However, simple fusion strategies such as addition or concatenation lack the flexibility to adaptively balance the contributions of each modality, often resulting in suboptimal feature integration.
To address this, the Local Adaptive Fusion module introduces an adaptive voxel-wise mechanism to dynamically combine camera and radar features.

LAF performs fine-grained fusion by first concatenating the 3D voxel features from the camera ($F_c$) and radar ($F_r$), followed by a lightweight convolutional network $\mathcal{F}_w$ \cite{wang2023openoccupancy} that predicts adaptive fusion weights $W_{laf} \in \mathbb{R}^{1 \times H \times W \times Z}$.
These weights modulate the contribution of each modality, enabling the network to emphasize the rich semantic information from camera data or the robust geometric cues from radar, depending on the spatial context.

The fusion process is defined as:
\begin{equation}
F_{laf} = W_{laf} \odot F_c + (1 - W_{laf}) \odot F_r,
\end{equation}
where
\begin{equation}
W_{laf} = \sigma(\mathcal{F}_w(\mathtt{Concat}[F_c, F_r])).
\end{equation}
Here $\sigma(\cdot)$ denotes the sigmoid activation function, which constrains the learned weights to the range $[0, 1]$, and $\odot$ indicates element-wise multiplication.
By enabling voxel-wise weighting, LAF achieves adaptive fusion while maintaining computational efficiency, offering flexible and robust integration of multi-modal features.

\textbf{Global Cross-Attention Fusion (GCF).}
Sensor discrepancies, such as differing sampling rates and transmission delays, can introduce spatio-temporal misalignments between radar and camera features, which may degrade the effectiveness of fusion.
To address this issue, we introduce the Global Cross-Attention Fusion module, which employs deformable attention with learnable sampling offsets to improve cross-modal alignment and integration, as illustrated in Fig.~\ref{fig:gcf}.

To reduce the computational cost of performing feature interaction directly in 3D voxel space, the input features are first projected into a 2D BEV representation:
\begin{equation}
F^{bev} = \mathtt{Linear}(\mathcal{H}(F^{3D})).
\end{equation}
Here, $\mathcal{H}(\cdot)$ denotes a height-to-channel (H2C) operator that flattens the vertical axis, and $\mathtt{Linear}$ is a projection layer that maps the reshaped tensor into the BEV space.
This produces BEV features $F^{bev} \in \mathbb{R}^{C' \times H \times W}$, where $C' = C \times Z$.

Unlike traditional attention mechanisms \cite{Transformer}, which use a single shared query for all inputs, the GCF module adopts a dual-stream deformable cross-attention design.
This structure allows modality-specific feature extraction while facilitating flexible and adaptive cross-modal interactions.
The fused BEV feature $F_{laf}^{bev}$, obtained from the Local Adaptive Fusion module, serves as the query input.
Using this semantically informed query provides a stronger prior than randomly initialized queries, accelerating convergence and improving alignment quality.

Each modality stream is equipped with a learnable positional encoding $P_{k \in \{c,r\}} \in \mathbb{R}^{C' \times H \times W}$, allowing the network to model spatial dependencies unique to camera ($k=c$) and radar ($k=r$) inputs.
Multi-head deformable attention (MDA) is applied to both streams using these encodings, with learnable offsets enabling spatially adaptive sampling and alignment.
The outputs are aggregated, normalized, and refined to form the final global fused representation. The overall GCF process is expressed as Eq. (\ref{EQ_F_m}), 
\begin{equation}
\label{EQ_F_m}
F_{m} = \mathtt{Conv}\left(\mathcal{H}'\left(\sum_{k \in \{c,r\}} \mathtt{MDA}\left(F_{laf}^{bev}, F_{k}^{bev}, P_k\right)\right) + F_{laf}\right)
\end{equation}
where $\mathtt{MDA}$ denotes multi-head deformable attention, $F_k^{bev}$ are the BEV features of modality $k$, and $\mathcal{H}'(\cdot)$ is the channel-to-height (C2H) operator that reshapes BEV features back into the 3D voxel space.
The final result is refined via a 3D convolution $\mathtt{Conv}(\cdot)$ and combined with the original LAF output through a residual connection.

\textbf{Temporal Fusion.}
Temporal fusion across multiple frames is crucial for accurate occupancy prediction, particularly in scenarios involving occlusions or fast motion.
Inspired by the approach in \cite{wang2024panoocc}, we introduce a Temporal Fusion module that effectively integrates temporal context from past frames, as illustrated in Fig.~\ref{fig:TAF}.

Given a sequence of 3D features $\{F_T, F_{T-1}, \cdots, F_{T-N}\} \in \mathbb{R}^{C \times H \times W \times Z}$, the fusion process begins by initializing a 3D reference point $F_{ini}$ at the current frame $T$.
To align historical features with the current frame, $F_{ini}$ is transformed to earlier timestamps using a coordinate transformation function $\mathcal{T}(\cdot)$ parameterized by the global pose $\mathcal{P}_{T \rightarrow (T-k)}$.
Aligned features from each previous frame are extracted via trilinear grid sampling $\mathcal{G}_t$, producing $F_{(T-k) \rightarrow T}$ as temporally aligned features:
\begin{equation}
F_{(T-k) \rightarrow T} = \mathcal{G}_t(\mathcal{T}(F_{ini}, \mathcal{P}_{T \rightarrow (T-k)})).
\end{equation}

After alignment, the set of 3D multi-modal features $\{F_T, F_{(T-1)\rightarrow T}, \cdots, F_{(T-N)\rightarrow T}\}$ is concatenated and passed through a refinement block to generate the final temporally enhanced feature $F$:
\begin{equation}
F = \mathtt{BottleNeck}(\mathtt{Concat}[F_T, \cdots, F_{(T-N)\rightarrow T}])
\end{equation}
Here, $\mathtt{BottleNeck}(\cdot)$ denotes a 3D convolutional network comprising convolution layers, batch normalization, and ReLU activations.
This module allows the network to leverage both current and historical context, improving occupancy prediction robustness in dynamic and occluded scenes.

\subsection{Occupany Head}
The occupancy prediction head consists of a series of linear layers that map the fused multi-modal voxel features to occupancy probability distributions.
To supervise training, we adopt a composite loss function designed to optimize both semantic accuracy and geometric consistency.

The primary loss component is the cross-entropy loss $\mathcal{L}_{ce}$, which provides voxel-wise supervision for occupancy classification.
To further improve segmentation boundary quality, we incorporate the Lovász-Softmax loss $\mathcal{L}_{lovasz}$ \cite{wang2023openoccupancy}.
Additionally, two scene-class affinity losses, a geometric affinity loss $\mathcal{L}_{scal}^{geo}$ and a semantic affinity loss $\mathcal{L}_{scal}^{sem}$, are included to encourage local consistency in geometric structure and semantic categories, as proposed in \cite{monoscene}.

The overall loss function is formulated as:
\begin{equation}
\mathcal{L} = \lambda_1\mathcal{L}_{ce} + \lambda_2\mathcal{L}_{lovasz} + \lambda_3\mathcal{L}_{scal}^{geo} + \lambda_4\mathcal{L}_{scal}^{sem}
\end{equation}
where the weights are set as $\lambda_1 = 1$, $\lambda_2 = 5$, $\lambda_3 = 1$, and $\lambda_4 = 1$.
%

\begin{figure*}[ht]
    \centering
    \includegraphics[width=0.95\linewidth]{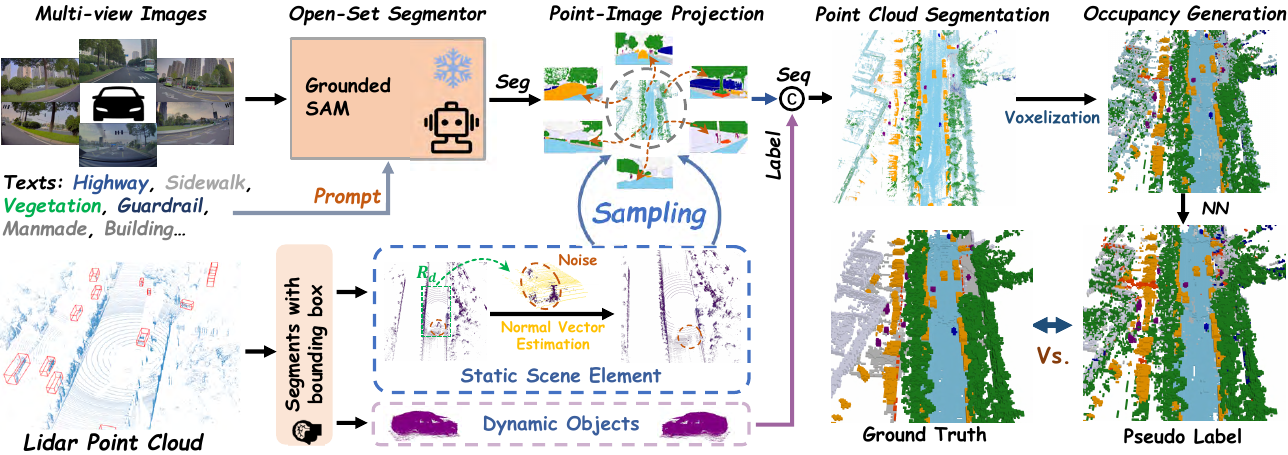}
    \caption{
    Pseudo-label generation pipeline. (a) Zero-shot multi-view semantic segmentation is performed using Grounded-SAM guided by fine-tuned text prompts. (b) LiDAR point clouds are separated into dynamic objects and static elements using 3D bounding boxes, followed by rain noise suppression via feature-based filtering within a coarse drivable region $R_d$, and semantic labeling through point-image projection. (c) Temporal enhancement aggregates sequential frames to improve semantic consistency. (d) Final occupancy pseudo-labels are produced through voxelization and a staged nearest-neighbor matching algorithm.
    }
    \label{fig:SemiOcc_Pipeline}
\end{figure*}

\subsection{Pseudo-Label Generation Approach}
Building on the rapid advances in large-scale image segmentation models, particularly those capable of zero-shot segmentation using text prompts, we propose a novel pipeline for generating high-quality pseudo-labels to reduce annotation costs in 3D occupancy prediction.
The overall framework is illustrated in Fig.\ref{fig:SemiOcc_Pipeline}, with implementation details provided in Algorithm\ref{alg:pseudo_label_pipeline}.

\textbf{Image Semantic Segmentation.}
We adopt Grounded-SAM \cite{ren2024grounded}, a vision-language segmentation model pretrained on over 10 million images, as our image segmentor.
To fully exploit its capabilities, we design a task-specific prompt set $\mathcal{T}$ containing semantically rich text queries tailored to common classes in driving scenes.
To improve segmentation diversity and robustness, multiple semantically similar prompts are assigned to each class.
Grounded-SAM is then applied to each image $I_k \in \mathcal{I}$, generating high-quality 2D semantic masks $I_k^{seg}$.

\textbf{LiDAR Semantic Segmentation.}
Following standard occupancy labeling procedures \cite{OmniHDScenes, surroundocc}, we use annotated 3D bounding boxes $\mathcal{B}$ to separate LiDAR points into dynamic objects $D_{obj}$ and static elements $S_{ele}$.
Instead of manually annotating static LiDAR points, we project them onto the image masks $I_k^{seg}$ to assign semantic labels.
To resolve inconsistencies arising from multi-view projections, we employ a confidence-based ranking strategy that selects the most reliable semantic prediction for each point, resulting in a unified static semantic point cloud $S^{seg}$.

Under adverse weather (e.g., rain), conventional radius- or voxel-based filters often fail to remove noise near vehicles, leading to degraded label quality, illustrated by the brown dashed box in Fig.\ref{fig:SemiOcc_Pipeline}.
To mitigate this, we propose a feature-based noise filtering approach.
First, a coarse drivable region $R_d$ is defined based on the ego vehicle's location and nearby 3D bounding boxes (see green dashed box in Fig.\ref{fig:SemiOcc_Pipeline}).
Within $R_d$, we estimate normal vectors to identify ground surfaces and remove non-ground points.
This produces a clean, semantically labeled static point cloud $S_{ele}^*$ suitable for pseudo-label generation.

\textbf{Temporal Enhancement.}
To reduce occlusion effects, we integrate temporal information from multiple frames.
Dynamic objects are aggregated by tracking IDs to produce $D_{seq}$, while static elements are spatially aligned based on pose $\mathcal{P}$ and aggregated in the global coordinate system to form the combined point cloud $\mathcal{L}^{seg}$.

\textbf{Occupancy Generation.}
We transform $\mathcal{L}^{seg}$ into the local frame, concatenate it with $D_{seq}$, and voxelize the resulting point cloud to produce $P^{seg}$ for pseudo-label generation.
Because dynamic object semantics are derived from manually annotated boxes and static element semantics from image segmentation, simple nearest-neighbor matching is insufficient.
To resolve this, we adopt a staged nearest-neighbor matching strategy: voxels containing dynamic points are matched to dynamic object labels, while the remaining voxels are matched to static points to generate the final pseudo-labels $\hat{\mathcal{O}}$.
\begin{algorithm}[t]
\caption{Pseudo-Label Generation Pipeline}
\label{alg:pseudo_label_pipeline}

\textbf{Input:} Multi-view images Sequence $\mathcal{I}$, Lidar Sequence $\mathcal{L}$, Text prompts $\mathcal{T}$, Poses $\mathcal{P}$, 3D bounding boxes $\mathcal{B}$.

\textbf{Output:} Occupancy pseudo-labels $\hat{\mathcal{O}}$ with semantics

\begin{algorithmic}
    \State \textcolor{magenta}{// Image Semantic Segmentation}
    \For{each image $I_k \in \mathcal{I}, k \in \{f,fl,fr,b,bl,br\}$}
        \State $I_k^{seg} \leftarrow \text{GroundedSAM}(I_k, \mathcal{T})$
    \EndFor
    
    \State \textcolor{magenta}{// LiDAR Semantic Segmentation}
    \For{each point cloud $P \in \mathcal{L}$}
        \State $D_{obj} \leftarrow \text{ExtractObject}(P, \mathcal{B})$
        \State $S_{ele} \leftarrow \text{Separate}(P, D_{obj})$
        \State $R_{d} \leftarrow \text{DefineDrivableRegion}(\text{ego pose}, \mathcal{B})$
        \State $S_{ele}^* \leftarrow \text{NormalEstimation}(S_{ele}, R_{d}) \triangleright \text{Noise Point}$ 
        \State $S^{seg} \leftarrow \text{Point-Image Projection}(S_{ele}^*, I_k^{seg})$
    \EndFor

    \State \textcolor{magenta}{// Temporal Enhancement}
    \State $D_{seq} \leftarrow \text{SequenceAgg}(D_{obj}, \text{ID})$
    \State $\mathcal{L}^{seg} \leftarrow \text{Concat}(\text{TransformToGlobal}(S^{seg}, \mathcal{P}))$

    \State \textcolor{magenta}{// Occupancy Generation}
    \State $P^{seg} \leftarrow \text{Concat}(\text{TransformToLocal}(\mathcal{L}^{seg}, \mathcal{P}), D_{seq})$
    \State $V_{seg} \leftarrow \text{Voxelization}(P^{seg})$
    \State $\hat{\mathcal{O}} \leftarrow \text{StagedNearestNeighbor}(V_{seg}, P^{seg})$
    
    \State \Return $\hat{\mathcal{O}}$
\end{algorithmic}
\end{algorithm}


\definecolor{ncar}{RGB}{255, 165, 0}
\definecolor{npedestrian}{RGB}{128, 0, 128}
\definecolor{nrider}{RGB}{0, 0, 200}
\definecolor{nlarge_vehicle}{RGB}{220, 220, 0}
\definecolor{ncycle}{RGB}{230 ,230 ,230}
\definecolor{nroad_obstacle}{RGB}{255, 69, 0}
\definecolor{ntraffic_fence}{RGB}{0, 0, 150}
\definecolor{ndriveable_surface}{RGB}{135 ,206 ,235}
\definecolor{nsidewalk}{RGB}{200, 200, 200}
\definecolor{nvegetation}{RGB}{34 ,139 ,34}
\definecolor{nmanmade}{RGB}{230, 230, 250}

\setlength{\tabcolsep}{1.4pt} 
\renewcommand{\arraystretch}{1} 

\begin{table*}[t]
\centering
\caption{
Quantitative results of 3D semantic occupancy prediction on the \textbf{OmniHD-Scenes test set}. The proposed MetaOcc framework achieves state-of-the-art performance, surpassing both camera-only and radar-camera fusion baselines. “C” and “R” indicate the use of camera and 4D radar modalities, respectively. The last eleven columns report per-class IoU scores across semantic categories. \textbf{Bold} and \underline{underlined} values indicate the best and second-best scores, respectively.
}
\label{occ_performance_omnihd}
\resizebox{1.0\textwidth}{!}{
\begin{tabular}{c|c|c|c|cc|ccccccccccc}
\hline
\noalign{\smallskip}
Methods     & Image Res. & Modality & Backbone & SC IoU  & mIoU  & \rotatebox{90}{\textcolor{ncar}{$\blacksquare$}car}    & \rotatebox{90}{\textcolor{npedestrian}{$\blacksquare$}pedestrian} & \rotatebox{90}{\textcolor{nrider}{$\blacksquare$}rider} & \rotatebox{90}{\textcolor{nlarge_vehicle}{$\blacksquare$}large vehicle} & \rotatebox{90}{\textcolor{ncycle}{$\blacksquare$}cycle} & \rotatebox{90}{\textcolor{nroad_obstacle}{$\blacksquare$}road obstacle} & \rotatebox{90}{\textcolor{ntraffic_fence}{$\blacksquare$}traffic fence} & \rotatebox{90}{\textcolor{ndriveable_surface}{$\blacksquare$}drive. surf.} & \rotatebox{90}{\textcolor{nsidewalk}{$\blacksquare$}sidewalk} & \rotatebox{90}{\textcolor{nvegetation}{$\blacksquare$}vegetation} & \rotatebox{90}{\textcolor{nmanmade}{$\blacksquare$}manmade} \\
\noalign{\smallskip}
\hline
\noalign{\smallskip}
C-CONet \cite{wang2023openoccupancy} (ICCV 2023)    & 544×960  & C    & ResNet50  & 25.69 & 13.42 & 20.03 & 3.51 & 11.71 & 16.62 & 0.79 & 1.14 & 22.75 & 33.57 & 14.82 & 17.73 & 4.93 \\
SurroundOcc \cite{surroundocc} (ICCV 2023) & 544×960  & C    & ResNet50  & 28.61 & 15.20 & 21.46 & 3.96 & 10.76 & 16.58 & 1.57 & 2.99 & 21.63 & 48.52 & 18.31 & 16.73 & 4.71 \\
BEVFormer \cite{bevformer} (ECCV 2022)  & 544×960  & C    & ResNet50  & 27.04 & 14.97 & 20.64 & 5.87 & 14.40 & 16.68 & 1.52 & 3.64 & 20.64 & 46.61 & 16.19 & 14.80 & 3.69 \\
BEVFormer-T \cite{bevformer} (ECCV 2022) & 544×960  & C    & ResNet50  & 28.42 & 16.23 & 22.73 & 5.45 & 14.70  & 18.21 & 3.09 & 3.87 & 21.54 & 48.15 & 17.58 & 17.77 & 5.48 \\
PanoOcc~\cite{wang2024panoocc} (CVPR 2024)          &544×960       & C      & ResNet50  &26.36  &15.20  &22.42  &5.91    &13.58  & 17.98 & 3.11 & 3.36 & 21.46 & \textbf{50.47} & 15.90 & 11.20 & 1.80   \\
\noalign{\smallskip}
\hline
\noalign{\smallskip}
BEVFormer \cite{bevformer} (ECCV 2022)  & 864×1536 & C    & ResNet101-DCN & 28.30 & 16.41 & 23.72 & 6.37 & 16.33 & 20.44 & 1.78 & 3.78 & 22.21 & 48.55 & 17.88 & 15.49 & 3.99 \\
BEVFormer-T \cite{bevformer} (ECCV 2022) & 864×1536 & C    & ResNet101-DCN & 29.74 & 17.49 & 24.90 & 6.48 & 16.45 & 21.49 & 2.87 & 4.62 & 22.51 & 49.92 & 18.59 & 18.53 & 5.96 \\
\noalign{\smallskip}
\hline
\noalign{\smallskip}
BEVFusion~\cite{BEVFusion} (NeurIPS 2022)   & 544×960  & C+R & ResNet50  & 27.02 & 16.24 & \underline{27.02} & 4.78 & 21.71 & 21.59 & 1.55 & 2.78 & 25.21 & 44.35 & 12.32 & 13.06 & 4.25 \\
M-CONet  \cite{wang2023openoccupancy}  (ICCV 2023)  & 544×960  & C+R & ResNet50  & 27.74 & 16.08 & 25.21 & 3.42 & 17.53 & 21.46 & 0.88 & 0.58 & 29.88 & 34.48 & 14.89 & 19.57 & 8.98 \\

OccFusion~\cite{ming2024occfusion} (TIV 2024)   & 544×960  & C+R & ResNet50  & 30.66 & 17.64 & 20.20 & 6.62 & 18.00 & 19.72 & 2.29 & 3.99 & 32.35 & 46.10 & 17.00 & 18.16 & 9.60 \\
TEOcc  \cite{lin2024teocc}  (ECAI 2024)  & 544×960  & C+R & ResNet50  & \underline{32.28} & 17.71 & 26.79 & 5.03 & 19.22 & 21.93 & 1.67 & 2.87 & 26.89 & 47.96 & 13.85 & \underline{21.12} & 7.52 \\

\noalign{\smallskip}
\hline
\noalign{\smallskip}
\textbf{MetaOcc-S (ours)}  & 544×960 & C+R & ResNet50 & 31.52 & \underline{20.92} & 26.87 & \underline{7.21} & \underline{23.44} & \textbf{23.55} & \underline{3.96} & \underline{5.92} & \underline{37.22} & 49.07 & \underline{19.41} & 20.98 & \underline{12.46} \\

\textbf{MetaOcc (ours)}  & 544×960 & C+R & ResNet50 & \textbf{32.75} & \textbf{21.73} & \textbf{27.52} & \textbf{8.84} & \textbf{24.06} & \underline{23.35} & \textbf{4.79} & \textbf{6.47} & \textbf{38.75} & \underline{50.29} & \textbf{19.91} & \textbf{21.88} & \textbf{13.22}  \\

\noalign{\smallskip}
\hline
\end{tabular}}
\label{tab:occ_performance}
\end{table*}
\setlength{\tabcolsep}{1.4pt}

\definecolor{ncar}{RGB}{255, 165, 0}
\definecolor{npedestrian}{RGB}{128, 0, 128}
\definecolor{nrider}{RGB}{0, 0, 200}
\definecolor{nlarge_vehicle}{RGB}{220, 220, 0}
\definecolor{ncycle}{RGB}{230 ,230 ,230}
\definecolor{nroad_obstacle}{RGB}{255, 69, 0}
\definecolor{ntraffic_fence}{RGB}{0, 0, 150}
\definecolor{ndriveable_surface}{RGB}{135 ,206 ,235}
\definecolor{nsidewalk}{RGB}{200, 200, 200}
\definecolor{nvegetation}{RGB}{34 ,139 ,34}
\definecolor{nmanmade}{RGB}{230, 230, 250}

\setlength{\tabcolsep}{1.4pt}  %
\renewcommand{\arraystretch}{1} 

\begin{table*}[t]
\centering
\caption{
Quantitative results of 3D semantic occupancy prediction on the \textbf{OmniHD-Scenes adverse scenario subset}. “C” and “R” indicate the use of camera and 4D radar modalities, respectively. The last eleven columns report per-class IoU scores across semantic categories. \textbf{Bold} and \underline{underlined} values indicate the best and second-best scores, respectively.
}
\label{occ_performance_omnihd_adverse}
\resizebox{1.0\textwidth}{!}{
\begin{tabular}{c|c|c|c|cc|ccccccccccc}
\hline
\noalign{\smallskip}
Methods     & Image Res. & Modality & Backbone & SC IoU  & mIoU  & \rotatebox{90}{\textcolor{ncar}{$\blacksquare$}car}    & \rotatebox{90}{\textcolor{npedestrian}{$\blacksquare$}pedestrian} & \rotatebox{90}{\textcolor{nrider}{$\blacksquare$}rider} & \rotatebox{90}{\textcolor{nlarge_vehicle}{$\blacksquare$}large vehicle} & \rotatebox{90}{\textcolor{ncycle}{$\blacksquare$}cycle} & \rotatebox{90}{\textcolor{nroad_obstacle}{$\blacksquare$}road obstacle} & \rotatebox{90}{\textcolor{ntraffic_fence}{$\blacksquare$}traffic fence} & \rotatebox{90}{\textcolor{ndriveable_surface}{$\blacksquare$}drive. surf.} & \rotatebox{90}{\textcolor{nsidewalk}{$\blacksquare$}sidewalk} & \rotatebox{90}{\textcolor{nvegetation}{$\blacksquare$}vegetation} & \rotatebox{90}{\textcolor{nmanmade}{$\blacksquare$}manmade} \\
\noalign{\smallskip}
\hline
\noalign{\smallskip}
C-CONet \cite{wang2023openoccupancy} (ICCV 2023)    & 544×960  & C    & ResNet50  & 24.68 & 12.32 & 19.62 & 3.03 & 10.03 & 15.00 & 1.08 & 0.51 & 16.18 & 31.83 & 15.23 & 17.48 & 5.47  \\
SurroundOcc \cite{surroundocc} (ICCV 2023) & 544×960  & C    & ResNet50  & 27.20 & 14.04 & 21.07 & 3.78 & 9.56  & 15.03 & 1.97 & 2.41 & 14.85 & 45.58 & 18.63 & 16.07 & 5.43  \\
BEVFormer \cite{bevformer} (ECCV 2022)  & 544×960  & C    & ResNet50  & 25.55 & 13.65 & 20.60 & 5.41 & 11.99 & 15.35 & 2.17 & 2.48 & 13.43 & 43.16 & 16.94 & 14.41 & 4.25  \\
BEVFormer-T \cite{bevformer} (ECCV 2022) & 544×960  & C    & ResNet50  & 27.08 & 14.93 & 22.68 & 4.66 & 12.27 & 17.23 & 4.11 & 2.59 & 14.87 & 44.88 & 17.23 & 17.33 & 6.41  \\
PanoOcc~\cite{wang2024panoocc} (CVPR 2024)          &544×960       & C      & ResNet50  & 24.02 & 14.02 & 25.02 & 4.11 & 10.52 & 20.17 & 3.63 & 3.35 & 11.59 & 43.83 & 15.64 & 12.76 & 3.59 \\
\noalign{\smallskip}
\hline
\noalign{\smallskip}
BEVFormer \cite{bevformer} (ECCV 2022)  & 864×1536 & C    & ResNet101-DCN & 26.66 & 15.03 & 23.52 & 5.85 & 13.61 & 19.62 & 2.62 & 2.28 & 14.59 & 45.21 & 18.55 & 14.82 & 4.70  \\
BEVFormer-T \cite{bevformer} (ECCV 2022) & 864×1536 & C    & ResNet101-DCN & 28.41 & 15.84 & 24.59 & 5.69 & 13.37 & 20.09 & 3.59 & 3.04 & 13.80 & \underline{46.46} & 18.74 & 18.22 & 6.62  \\
\noalign{\smallskip}
\hline
\noalign{\smallskip}
BEVFusion~\cite{BEVFusion} (NeurIPS 2022)   & 544×960  & C+R & ResNet50  & 25.32 & 15.36 & \underline{27.17} & 3.58 & 19.17 & 22.62 & 2.31 & 1.55 & 21.72 & 40.93 & 12.54 & 12.64 & 4.69  \\
M-CONet  \cite{wang2023openoccupancy}  (ICCV 2023)  & 544×960  & C+R & ResNet50  & 26.73 & 15.30 & 25.37 & 3.07 & 16.13 & 20.97 & 1.41 & 0.25 & 23.95 & 32.65 & 15.47 & 19.75 & 9.21  \\

OccFusion~\cite{ming2024occfusion} (TIV 2024)   & 544×960  & C+R & ResNet50  & 29.54 & 17.27 & 20.34 & \underline{7.14} & 17.84 & 21.62 & 3.21 & 3.50 & 25.99 & 43.74 & 18.39 & 18.01 & 10.17 \\
TEOcc  \cite{lin2024teocc}  (ECAI 2024)  & 544×960  & C+R & ResNet50  & 29.96 & 16.30 & 26.34 & 3.97 & 16.96 & 21.82 & 2.58 & 1.47 & 20.37 & 44.23 & 13.44 & 20.01 & 8.07  \\

\noalign{\smallskip}
\hline
\noalign{\smallskip}
\textbf{MetaOcc-S (ours)}  & 544×960 & C+R & ResNet50 & \underline{30.24} & \underline{19.85} & 26.91 & 6.56 & \underline{21.06} & \textbf{24.27} & \underline{5.43} & \underline{4.22} & \underline{30.95} & 45.65 & \underline{19.85} & \underline{20.51} & \underline{12.96} \\

\textbf{MetaOcc (ours)}  & 544×960 & C+R & ResNet50 & \textbf{31.43} & \textbf{20.62} & \textbf{27.64} & \textbf{8.24} & \textbf{22.03} & \underline{23.94} & \textbf{6.21} & \textbf{4.54} & \textbf{31.72} & \textbf{46.74} & \textbf{20.34} & \textbf{21.56} & \textbf{13.85}  \\

\noalign{\smallskip}
\hline
\end{tabular}}
\label{tab:occ_performance_adverse}
\end{table*}
\setlength{\tabcolsep}{1.4pt}

\section{Experiments}
\label{sec:experiments}
\subsection{Implementation Details}
\textbf{Dataset.}
The \textit{OmniHD-Scenes} dataset \cite{OmniHDScenes} is one of the few publicly available multi-modal datasets that provide synchronized data from six surround-view 4D millimeter-wave radars and cameras.
It includes typical urban driving scenarios and consists of 200 annotated clips, each approximately 30 seconds in length, totaling 11,921 frames with 3D occupancy labels spanning 11 semantic classes.
The dataset is divided into 8,321 training frames and 3,600 test frames.

To evaluate the effectiveness of our proposed pseudo-label generation method, we construct a semi-supervised variant, \textit{OmniHD-SemiOcc}, by replacing 0\%, 50\%, and 100\% of the ground-truth labels in the training set with generated pseudo-labels.
We train multiple models on these semi-supervised variants and evaluate their performance on the original \textit{OmniHD-Scenes} test set to validate the utility of our labeling strategy.

To examine the generalization ability of the proposed MetaOcc framework beyond 4D radar, we further evaluate it under traditional radar-camera fusion settings using the \textit{SurroundOcc-nuScenes} dataset \cite{nuscenes, surroundocc}.
This dataset contains over 1,000 autonomous driving scenes, equipped with multi-view cameras and conventional millimeter-wave radars, and serves as a widely used benchmark for radar-camera fusion tasks.

\textbf{Metrics.} 
To evaluate the performance of different benchmarks, including \textit{OmniHD-Scenes}, \textit{OmniHD-SemiOcc}, and \textit{SurroundOcc-nuScenes}, we adopt a combination of semantic and geometric metrics.
In this paper, unless otherwise specified, Intersection over Union (IoU) refers to semantic IoU, which is computed per semantic class to measure the voxel-wise segmentation accuracy.
The mean IoU (mIoU) aggregates these scores over all semantic classes to provide a comprehensive evaluation of segmentation performance \cite{monoscene}.

For geometric evaluation, we use Scene Completion IoU (SC IoU), which assesses voxel-wise occupancy by distinguishing between free and occupied space.
SC IoU is treated separately from semantic metrics and is particularly useful for measuring geometric completeness and structural accuracy.

The semantic IoU and mIoU metrics are defined as:
\begin{equation}
\mathrm{IoU} = \frac{TP}{TP + FP + FN}
\end{equation}
\begin{equation}
\mathrm{mIoU} = \frac{1}{C} \sum_{i=1}^{C} \frac{TP_i}{TP_i + FP_i + FN_i}
\end{equation}
where $TP$, $FP$, and $FN$ represent the number of true positives, false positives, and false negatives, respectively, and $C$ is the total number of semantic classes.
SC IoU is computed over binary occupancy predictions, measuring voxel-level agreement between predicted and ground-truth free/occupied space.

\textbf{Network Settings.}
For the \textit{OmniHD-Scenes} dataset, the prediction range is defined as $(-60, 60)$ m, $(-40, 40)$ m, and $(-3, 5)$ m along the $X$-, $Y$-, and $Z$-axes, respectively.
A voxel size of 0.5 m is used, resulting in a spatial resolution of $160 \times 240 \times 16$.
Multi-view images are resized to $544 \times 960$ and encoded using ResNet-50 \cite{he2016resnet} as the 2D image backbone.
Each 4D radar frame includes spatial coordinates $(x, y, z)$, Doppler velocity components $(v_x, v_y)$, radar amplitude, signal-to-noise ratio (SNR), and timestamps.

For the \textit{SurroundOcc-nuScenes} dataset, the prediction range is set to $(-50, 50)$ m along both the $X$- and $Y$-axes and $(-3, 5)$ m along the $Z$-axis.
The voxel size remains at 0.5 m, yielding a spatial resolution of $200 \times 200 \times 16$.
To ensure a fair comparison, the radar preprocessing pipeline follows the settings used in OccFusion \cite{ming2024occfusion}, and ResNet101-DCN \cite{he2016resnet} is adopted as the 2D image backbone.

\begin{figure*}[ht]
    \centering
    \includegraphics[width=\linewidth]{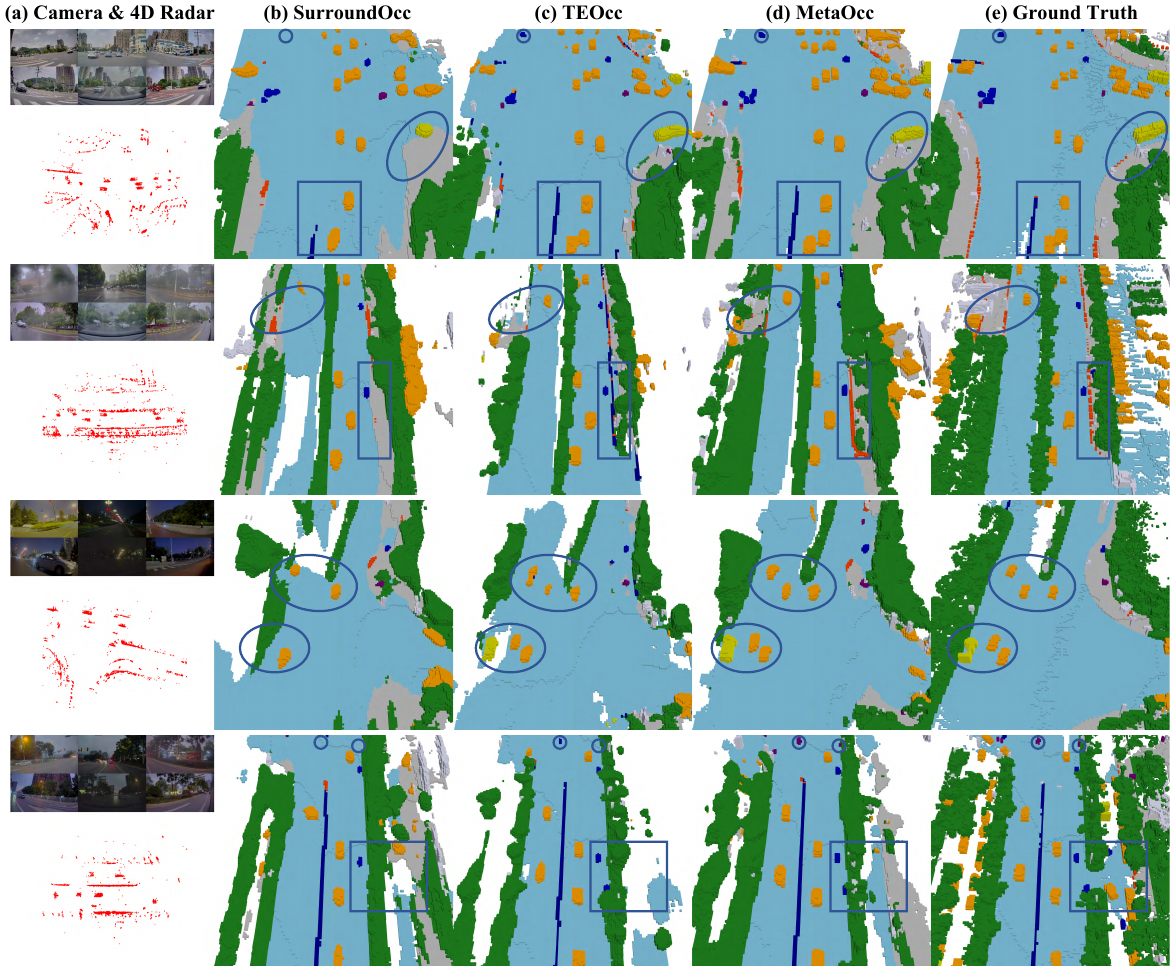}
    \caption{
    \textbf{Qualitative results across diverse scenes on the OmniHD-Scenes dataset}. (a) presents the multi-view camera images and 4D radar inputs. (b)–(e) show the 3D occupancy predictions from SurroundOcc, TEOcc, and MetaOcc, alongside the ground truth. MetaOcc demonstrates consistently superior spatial and semantic accuracy across a variety of complex scenarios.
    }
\label{fig:vis_omnid}
\end{figure*}

\textbf{Training Details.}
The proposed model is implemented using the MMDetection3D framework~\cite{contributors2020mmdetection3d} and trained on NVIDIA GeForce RTX 3090 GPUs with a batch size of 1.
The 2D image backbone in the camera stream is initialized with weights pre-trained on FCOS3D~\cite{wang2021fcos3d}.
An end-to-end training strategy is employed using the AdamW optimizer, with an initial learning rate 2e-4.
The model is trained for 12 epochs on the \textit{OmniHD-Scenes} dataset and for 24 epochs on the \textit{SurroundOcc-nuScenes} dataset.

\subsection{Main Results}
\textbf{Full Supervision Results on OmniHD-Scenes.}
We evaluate the proposed MetaOcc and its single-frame variant, MetaOcc-S, on the \textit{OmniHD-Scenes} test set under full supervision.
As shown in Table~\ref{occ_performance_omnihd}, MetaOcc achieves state-of-the-art performance, attaining 32.75 SC IoU and 21.73 mIoU.
This surpasses previous multi-modal approaches, including OccFusion (+2.09 SC IoU, +4.09 mIoU) and TEOcc (+0.47 SC IoU, +4.02 mIoU), demonstrating the effectiveness of our fusion and radar encoding strategies.

Vision-based perception methods often struggle with small-scale object detection, such as riders and bicycles, due to occlusion and the limited spatial context available in 2D features.
Radar, by contrast, offers complementary advantages, including long-range sensing and radial velocity measurements.
Additionally, the diffraction characteristics of millimeter-wave signals enable radar to partially penetrate visual obstructions, improving perception in occluded scenes.
By leveraging the Radar Height Self-Attention module, MetaOcc effectively enhances spatial encoding of sparse 4D radar point clouds, leading to measurable improvements in class-wise accuracy.
In particular, MetaOcc improves the IoU for the rider and cycle categories by 4.84 and 1.42 points, respectively, compared to TEOcc.

We also evaluate the robustness of MetaOcc under challenging conditions such as rain and low-light environments using the adverse scenario subset.
As shown in Table~\ref{occ_performance_omnihd_adverse}, MetaOcc maintains strong performance, achieving 31.43 SC IoU and 20.62 mIoU on this subset.
This resilience is attributed to the Hierarchical Multi-scale Multi-modal Fusion module, which adaptively balances cross-modal inputs and effectively integrates temporal information.
Qualitative results, as illustrated in Fig.~\ref{fig:vis_omnid}, further demonstrate MetaOcc's superior ability to capture fine-grained scene structure and object boundaries, even in visually degraded scenarios.

\definecolor{nbarrier}{RGB}{255, 120, 50}
\definecolor{nbicycle}{RGB}{255, 192, 203}
\definecolor{nbus}{RGB}{255, 255, 0}
\definecolor{ncar}{RGB}{0, 150, 245}
\definecolor{nconstruct}{RGB}{0, 255, 255}
\definecolor{nmotor}{RGB}{200, 180, 0}
\definecolor{npedestrian}{RGB}{255, 0, 0}
\definecolor{ntraffic}{RGB}{255, 240, 150}
\definecolor{ntrailer}{RGB}{135, 60, 0}
\definecolor{ntruck}{RGB}{160, 32, 240}
\definecolor{ndriveable}{RGB}{255, 0, 255}
\definecolor{nother}{RGB}{139, 137, 137}
\definecolor{nsidewalk}{RGB}{75, 0, 75}
\definecolor{nterrain}{RGB}{150, 240, 80}
\definecolor{nmanmade}{RGB}{213, 213, 213}
\definecolor{nvegetation}{RGB}{0, 175, 0}

\setlength{\tabcolsep}{1.4pt}  
\renewcommand{\arraystretch}{1}

\begin{table*}[htop]
\centering
\caption{
3D semantic occupancy prediction results on the \textbf{SurroundOcc-nuScenes validation set}. The proposed method achieves state-of-the-art performance, surpassing both camera-only and radar-camera fusion baselines. “C” and “R” indicate camera and radar modalities, respectively. All models are trained using dense occupancy annotations provided by SurroundOcc~\cite{surroundocc}.
}
\label{tab:nuscenes_surroundocc_val}
\resizebox{1.0\textwidth}{!}{
\begin{tabular}{c|c|c|c|cc|cccccccccccccccc}
    \toprule
    Method & Backbone & Input Modality & Params & IoU & mIoU 
    & \rotatebox{90}{\textcolor{nbarrier}{$\blacksquare$} barrier}
    & \rotatebox{90}{\textcolor{nbicycle}{$\blacksquare$} bicycle}
    & \rotatebox{90}{\textcolor{nbus}{$\blacksquare$} bus}
    & \rotatebox{90}{\textcolor{ncar}{$\blacksquare$} car}
    & \rotatebox{90}{\textcolor{nconstruct}{$\blacksquare$} const. veh.}
    & \rotatebox{90}{\textcolor{nmotor}{$\blacksquare$} motorcycle}
    & \rotatebox{90}{\textcolor{npedestrian}{$\blacksquare$} pedestrian}
    & \rotatebox{90}{\textcolor{ntraffic}{$\blacksquare$} traffic cone}
    & \rotatebox{90}{\textcolor{ntrailer}{$\blacksquare$} trailer}
    & \rotatebox{90}{\textcolor{ntruck}{$\blacksquare$} truck}
    & \rotatebox{90}{\textcolor{ndriveable}{$\blacksquare$} drive. suf.}
    & \rotatebox{90}{\textcolor{nother}{$\blacksquare$} other flat}
    & \rotatebox{90}{\textcolor{nsidewalk}{$\blacksquare$} sidewalk}
    & \rotatebox{90}{\textcolor{nterrain}{$\blacksquare$} terrain}
    & \rotatebox{90}{\textcolor{nmanmade}{$\blacksquare$} manmade}
    & \rotatebox{90}{\textcolor{nvegetation}{$\blacksquare$} vegetation}
    \\
    \midrule
    BEVFormer \cite{bevformer} & ResNet101-DCN & C & 59M & 30.50 & 16.75 & 14.22 & 6.58 & 23.46 & 28.28 & 8.66 & 10.77 & 6.64 & 4.05 & 11.20 & 17.78 & 37.28 & 18.00 & 22.88 & 22.17 & 13.80 & 22.21 \\
    Atlas \cite{atlas} & - & C & - & 28.66 & 15.00 & 10.64 & 5.68 & 19.66 & 24.94 & 8.90 & 8.84 & 6.47 & 3.28 & 10.42 & 16.21 & 34.86 & 15.46 & 21.89 & 20.95 & 11.21 & 20.54 \\
    TPVFormer \cite{TPVFormer} & ResNet101-DCN & C & 69M & 30.86 & 17.10 & 15.96 & 5.31 & 23.86 & 27.32 & 9.79 & 8.74 & 7.09 & 5.20 & 10.97 & 19.22 & 38.87 & 21.25 & 24.26 & 23.15 & 11.73 & 20.81 \\
    C-CONet \cite{wang2023openoccupancy} & ResNet101 & C & 118M & 26.10 & 18.40 & 18.60 & 10.00 & 26.40 & 27.40 & 8.60 & 15.70 & 13.30 & 9.70 & 10.90 & 20.20 & 33.00 & 20.70 & 21.40 & 21.80 & 14.70 & 21.30 \\
    InverseMatrixVT3D \cite{ming2024inversematrixvt3d} & ResNet101-DCN & C & 67M & 
    30.03 & 18.88 & 18.39 & 12.46 & 26.30 & 29.11 & 11.00 & 15.74 & 14.78 & 11.38 & 13.31 & 21.61 & 36.30 & 19.97 & 21.26 & 20.43 & 11.49 & 18.47 \\
    SurroundOcc \cite{surroundocc} & ResNet101-DCN & C & 180M & 31.49 & 20.30 & 20.59 & 11.68 & 28.06 & 30.86 & 10.70 & 15.14 & 14.09 & 12.06 & 14.38 & 22.26 & 37.29 & 23.70 & 24.49 & 22.77 & 14.89 & 21.86 \\
    OccFormer \cite{zhang2023occformer} & ResNet101-DCN & C & 169M & 31.39 & 19.03 & 18.65 & 10.41 & 23.92 & 30.29 & 10.31 & 14.19 & 13.59 & 10.13 & 12.49 & 20.77 & 38.78 & 19.79 & 24.19 & 22.21 & 13.48 & 21.35 \\
    FB-Occ \cite{fbocc} & ResNet101 & C & - & 31.50 & 19.60 & 20.60 & 11.30 & 26.90 & 29.80 & 10.40 & 13.60 & 13.70 & 11.40 & 11.50 & 20.60 & 38.20 & 21.50 & 24.60 & 22.70 & 14.80 & 21.60 \\
    RenderOcc \cite{pan2024renderocc} & ResNet101 & C & 122M & 29.20 & 19.00 & 19.70 & 11.20 & 28.10 & 28.20 & 9.80 & 14.70 & 11.80 & 11.90 & 13.10 & 20.10 & 33.20 & 21.30 & 22.60 & 22.30 & 15.30 & 20.90 \\
    GaussianFormer \cite{gaussianformer} & ResNet101-DCN & C & - & 29.83 & 19.10 & 19.52 & 11.26 & 26.11 & 29.78 & 10.47 & 13.83 & 12.58 & 8.67 & 12.74 & 21.57 & 39.63 & 23.28 & 24.46 & 22.99 & 9.59 & 19.12 \\
    Co-Occ \cite{co-occ} & ResNet101 & C & 218M & 30.00 & 20.30 & \textbf{22.50} & 11.20 & \textbf{28.60} & 29.50 & 9.90 & 15.80 & 13.50 & 8.70 & 13.60 & 22.20 & 34.90 & 23.10 & 24.20 & \textbf{24.10} & \textbf{18.00} & \textbf{24.80} \\
    GaussianFormer2 \cite{gaussianformer2} & ResNet101-DCN & C & - & 31.14 & 20.36 & 19.93 & 12.99 & 28.15 & 30.82 & 10.97 & 16.54 & 13.23 & 10.56 & 13.39 & 22.20 & \textbf{39.71} & 23.65 & \textbf{25.43} & 23.68 & 12.96 & 21.51 \\

    Inverse++ \cite{ming2025inverse++} & ResNet101-DCN & C & 137M & 31.73 & 20.91 & 20.90 & 13.27 & 28.40 & 31.37 & 11.90 & 17.76 & 15.39 & 13.49 & 13.32 & 23.19 & 39.37 & 22.85 & 25.27 & 23.68 & 13.43 & 20.98 \\
    \noalign{\smallskip}
    \hline
    \noalign{\smallskip}
    OccFusion \cite{ming2024occfusion} & ResNet101-DCN & C+R & 93M & 32.90 & 20.73 & 20.46 & 13.98 & 27.99 & 31.52 & \textbf{13.68} & 18.45 & 15.79 & 13.05 & 13.94 & 23.84 & 37.85 & 19.60 & 22.41 & 21.20 & 16.16 & 21.81 \\
    \textbf{MetaOcc-S (ours)} & ResNet101-DCN & C+R & 69M & \textbf{34.06} & \textbf{21.97} & 21.97 & \textbf{14.23} & 28.19 & \textbf{33.29} & 13.55 & \textbf{18.85} & \textbf{16.50} & \textbf{13.58} & \textbf{14.39} & \textbf{24.78} & 39.55 & \textbf{24.36} & 24.65 & 22.41 & 17.81 & 23.44 \\
    \bottomrule
\end{tabular}}
\end{table*}
\setlength{\tabcolsep}{1.4pt}

\textbf{Full Supervision Results on SurroundOcc-nuScenes.}
To further evaluate the generalization and adaptability of our method across different radar hardware configurations, we transfer the single-frame variant MetaOcc-S to the \textit{SurroundOcc-nuScenes} dataset, which utilizes conventional surround-view radar sensors.

As shown in Table~\ref{tab:nuscenes_surroundocc_val}, MetaOcc-S achieves 34.06 SC IoU and 21.97 mIoU on the validation set, outperforming the strong camera-only baseline GaussianFormer2 by +2.92 SC IoU and +1.61 mIoU, and exceeding the radar-camera fusion method OccFusion by +1.16 SC IoU and +1.24 mIoU.
With only 69 million parameters, MetaOcc-S offers a favorable balance between accuracy and model complexity, confirming its robustness and efficiency.
These results demonstrate the model’s capacity to generalize across sensor domains and highlight its potential for practical deployment in real-world autonomous systems.

\definecolor{nbarrier}{RGB}{255, 120, 50}
\definecolor{nbicycle}{RGB}{255, 192, 203}
\definecolor{nbus}{RGB}{255, 255, 0}
\definecolor{ncar}{RGB}{0, 150, 245}
\definecolor{nconstruct}{RGB}{0, 255, 255}
\definecolor{nmotor}{RGB}{200, 180, 0}
\definecolor{npedestrian}{RGB}{255, 0, 0}
\definecolor{ntraffic}{RGB}{255, 240, 150}
\definecolor{ntrailer}{RGB}{135, 60, 0}
\definecolor{ntruck}{RGB}{160, 32, 240}
\definecolor{ndriveable}{RGB}{255, 0, 255}
\definecolor{nother}{RGB}{139, 137, 137}
\definecolor{nsidewalk}{RGB}{75, 0, 75}
\definecolor{nterrain}{RGB}{150, 240, 80}
\definecolor{nmanmade}{RGB}{213, 213, 213}
\definecolor{nvegetation}{RGB}{0, 175, 0}

\setlength{\tabcolsep}{1.4pt}  
\renewcommand{\arraystretch}{1}

\begin{table*}[ht]
\centering
\caption{
3D semantic occupancy prediction results on the \textbf{SurroundOcc-nuScenes validation rainy scenario subset}. “C” and “R” denote camera and radar modalities, respectively. All models are trained using dense occupancy annotations provided by SurroundOcc~\cite{surroundocc}.
}
\label{tab:nuscenes_surroundocc_val_rainy}
\resizebox{1.0\textwidth}{!}{
\begin{tabular}{c|c|c|cc|cccccccccccccccc}
    \toprule
    Method & Backbone & Input Modality & IoU & mIoU 
    & \rotatebox{90}{\textcolor{nbarrier}{$\blacksquare$} barrier}
    & \rotatebox{90}{\textcolor{nbicycle}{$\blacksquare$} bicycle}
    & \rotatebox{90}{\textcolor{nbus}{$\blacksquare$} bus}
    & \rotatebox{90}{\textcolor{ncar}{$\blacksquare$} car}
    & \rotatebox{90}{\textcolor{nconstruct}{$\blacksquare$} const. veh.}
    & \rotatebox{90}{\textcolor{nmotor}{$\blacksquare$} motorcycle}
    & \rotatebox{90}{\textcolor{npedestrian}{$\blacksquare$} pedestrian}
    & \rotatebox{90}{\textcolor{ntraffic}{$\blacksquare$} traffic cone}
    & \rotatebox{90}{\textcolor{ntrailer}{$\blacksquare$} trailer}
    & \rotatebox{90}{\textcolor{ntruck}{$\blacksquare$} truck}
    & \rotatebox{90}{\textcolor{ndriveable}{$\blacksquare$} drive. suf.}
    & \rotatebox{90}{\textcolor{nother}{$\blacksquare$} other flat}
    & \rotatebox{90}{\textcolor{nsidewalk}{$\blacksquare$} sidewalk}
    & \rotatebox{90}{\textcolor{nterrain}{$\blacksquare$} terrain}
    & \rotatebox{90}{\textcolor{nmanmade}{$\blacksquare$} manmade}
    & \rotatebox{90}{\textcolor{nvegetation}{$\blacksquare$} vegetation}
    \\
    \midrule
    InverseMatrixVT3D \cite{ming2024inversematrixvt3d} & ResNet101-DCN & C & 29.72 & 18.99 & 18.55 & 14.29 & 22.28 & 30.02 & 10.19 & 15.20 & 10.03 & 9.71 & 13.28 & 20.98 & 37.18 & 23.47 & 27.74 & 17.46 & 10.36 & 23.13 \\
    GaussianFormer \cite{gaussianformer} & ResNet101-DCN & C & 27.37 & 16.96 & 18.16 & 9.58 & 21.09 & 26.83 & 8.04 & 10.13 & 7.80 & 5.84 & 12.66 & 18.24 & 35.53 & 18.51 & 27.79 & 19.23 & 11.04 & 20.85 \\
    Co-Occ \cite{co-occ} & ResNet101 & C & 28.90 & 19.70 & 22.10 & \textbf{17.60} & 26.30 & 30.80 & 10.90 & 9.90 & 8.20 & 9.70 & 11.40 & 19.30 & 39.00 & 22.20 & \textbf{32.60} & \textbf{23.00} & 11.50 & 21.30 \\
    GaussianFormer2 \cite{gaussianformer2} & ResNet101-DCN & C & 31.14 & 20.36 & 19.84 & 13.52 & \textbf{26.89} & 31.65 & 10.82 & 15.16 & 9.04 & 8.41 & 13.72 & 21.84 & \textbf{40.51} & 24.57 & 32.21 & 20.65 & 12.64 & 24.33 \\
    SurroundOcc \cite{surroundocc} & ResNet101-DCN & C & 30.57 & 19.85 & 21.40 & 12.75 & 25.49 & 31.31 & 11.39 & 12.65 & 8.94 & 9.48 & 14.51 & 21.52 & 35.34 & \textbf{25.32} & 29.89 & 18.37 & 14.44 & 24.78 \\
    Inverse++ \cite{ming2025inverse++}&  ResNet101-DCN & C & 31.32 & 20.66 & \textbf{22.52} & 13.79 & 25.49 & 31.80 & \textbf{11.70} & 16.72 & \textbf{11.14} & 10.12 & 12.29 & 22.25 & 38.78 & 23.93 & 31.62 & 21.14 & 12.65 & 24.61 \\
    \noalign{\smallskip}
    \hline
    \noalign{\smallskip}
    OccFusion \cite{ming2024occfusion}& ResNet101-DCN & C+R & 33.75 & 20.78 & 20.14 & 16.33 & 26.37 & 32.39 & 11.56 & \textbf{17.08} & \textbf{11.14} & 10.54 & 13.61 & \textbf{22.42} & 37.50 & 22.79 & 29.50 & 17.58 & 17.06 & 26.49 \\
    \textbf{MetaOcc-S (ours)} & ResNet101-DCN & C+R & \textbf{33.81} & \textbf{21.51} & 21.95 & 15.74 & 26.26 & \textbf{34.06} & 11.30 & 12.74 & 10.99 & \textbf{10.65} & \textbf{14.92} & 22.26 & 39.72 & 26.48 & 31.41 & 19.63 & \textbf{18.55} & \textbf{27.54} \\
    \bottomrule
\end{tabular}}
\end{table*}
\setlength{\tabcolsep}{1.4pt}

\definecolor{nbarrier}{RGB}{255, 120, 50}
\definecolor{nbicycle}{RGB}{255, 192, 203}
\definecolor{nbus}{RGB}{255, 255, 0}
\definecolor{ncar}{RGB}{0, 150, 245}
\definecolor{nconstruct}{RGB}{0, 255, 255}
\definecolor{nmotor}{RGB}{200, 180, 0}
\definecolor{npedestrian}{RGB}{255, 0, 0}
\definecolor{ntraffic}{RGB}{255, 240, 150}
\definecolor{ntrailer}{RGB}{135, 60, 0}
\definecolor{ntruck}{RGB}{160, 32, 240}
\definecolor{ndriveable}{RGB}{255, 0, 255}
\definecolor{nother}{RGB}{139, 137, 137}
\definecolor{nsidewalk}{RGB}{75, 0, 75}
\definecolor{nterrain}{RGB}{150, 240, 80}
\definecolor{nmanmade}{RGB}{213, 213, 213}
\definecolor{nvegetation}{RGB}{0, 175, 0}

\setlength{\tabcolsep}{1.4pt}  
\renewcommand{\arraystretch}{1}

\begin{table*}[ht]
\centering
\caption{
3D semantic occupancy prediction results on the \textbf{SurroundOcc-nuScenes validation night scenario subset}. “C” and “R” denote camera and radar modalities, respectively. All models are trained using dense occupancy annotations provided by SurroundOcc~\cite{surroundocc}.
}
\label{tab:nuscenes_surroundocc_val_night}
\resizebox{1.0\textwidth}{!}{
\begin{tabular}{c|c|c|cc|cccccccccccccccc}
    \toprule
    Method & Backbone & Input Modality & IoU & mIoU 
    & \rotatebox{90}{\textcolor{nbarrier}{$\blacksquare$} barrier}
    & \rotatebox{90}{\textcolor{nbicycle}{$\blacksquare$} bicycle}
    & \rotatebox{90}{\textcolor{nbus}{$\blacksquare$} bus}
    & \rotatebox{90}{\textcolor{ncar}{$\blacksquare$} car}
    & \rotatebox{90}{\textcolor{nconstruct}{$\blacksquare$} const. veh.}
    & \rotatebox{90}{\textcolor{nmotor}{$\blacksquare$} motorcycle}
    & \rotatebox{90}{\textcolor{npedestrian}{$\blacksquare$} pedestrian}
    & \rotatebox{90}{\textcolor{ntraffic}{$\blacksquare$} traffic cone}
    & \rotatebox{90}{\textcolor{ntrailer}{$\blacksquare$} trailer}
    & \rotatebox{90}{\textcolor{ntruck}{$\blacksquare$} truck}
    & \rotatebox{90}{\textcolor{ndriveable}{$\blacksquare$} drive. suf.}
    & \rotatebox{90}{\textcolor{nother}{$\blacksquare$} other flat}
    & \rotatebox{90}{\textcolor{nsidewalk}{$\blacksquare$} sidewalk}
    & \rotatebox{90}{\textcolor{nterrain}{$\blacksquare$} terrain}
    & \rotatebox{90}{\textcolor{nmanmade}{$\blacksquare$} manmade}
    & \rotatebox{90}{\textcolor{nvegetation}{$\blacksquare$} vegetation}
    \\
    \midrule
    InverseMatrixVT3D \cite{ming2024inversematrixvt3d} & ResNet101-DCN & C & 22.48 & 9.99 & 10.40 & 12.03 & 0.00 & 29.94 & 0.00 & 9.92 & 4.88 & \textbf{0.91} & 0.00 & 17.79 & 29.10 & 2.37 & 10.80 & 9.40 & 8.68 & 13.57 \\
    GaussianFormer \cite{gaussianformer} & ResNet101-DCN & C & 20.30 & 9.07 & 6.11 & 8.70 & 0.00 & 25.75 & 0.00 & 10.44 & 2.85 & 0.55 & 0.00 & 17.26 & 30.65 & \textbf{2.95} & 12.53 & 9.94 & 6.65 & 10.71 \\
    Co-Occ \cite{co-occ} & ResNet101 & C & 18.90 & 9.40 & 4.50 & 9.30 & 0.00 & 29.50 & 0.00 & 8.40 & 3.50 & 0.00 & 0.00 & 15.10 & 29.40 & 0.60 & 12.40 & 11.50 & 10.70 & 15.50 \\
    GaussianFormer2 \cite{gaussianformer2} & ResNet101-DCN & C & 21.19 & 10.14 & 5.25 & 9.29 & 0.00 & 29.33 & 0.00 & 13.65 & 5.80 & 0.90 & 0.00 & 20.22 & 31.80 & 1.94 & \textbf{14.83} & 10.48 & 5.96 & 12.72 \\
    SurroundOcc \cite{surroundocc} & ResNet101-DCN & C & 24.38 & 10.80 & \textbf{10.55} & \textbf{14.60} & 0.00 & 31.05 & 0.00 & 8.26 & 5.37 & 0.58 & 0.00 & 18.75 & 30.72 & 2.74 & 12.39 & 11.53 & 10.52 & 15.77 \\
    Inverse++ \cite{ming2025inverse++} & ResNet101-DCN & C & 23.70 & 10.93 & 8.87 & 10.19 & 0.00 & 32.62 & 0.00 & 11.77 & 7.46 & 0.72 & 0.00 & \textbf{22.20} & 32.95 & 2.15 & 13.01 & 9.79 & 8.61 & 14.48 \\
    \noalign{\smallskip}
    \hline
    \noalign{\smallskip}
    OccFusion \cite{ming2024occfusion}& ResNet101-DCN & C+R & 27.09 & 11.13 & 10.78 & 12.77 & 0.00 & 33.50 & 0.00 & 12.72 & 4.91 & 0.61 & 0.00 & 19.97 & 29.51 & 0.94 & 12.15 & 10.72 & 11.81 & 17.72 \\
    \textbf{MetaOcc-S (ours)} & ResNet101-DCN & C+R & \textbf{28.35} & \textbf{12.29} & 7.91  & 12.37 & 0.00  & \textbf{36.19} & 0.00  & \textbf{13.89} & \textbf{9.40}  & 0.88  & 0.00  & 19.71 & \textbf{33.78} & 2.82  & 14.51 & \textbf{12.94} & \textbf{12.32} & \textbf{19.86} \\
    \bottomrule
\end{tabular}}
\end{table*}
\setlength{\tabcolsep}{1.4pt}
We further analyze the model’s robustness under adverse environmental conditions using rainy and night subsets of the \textit{SurroundOcc-nuScenes} dataset, as reported in Table~\ref{tab:nuscenes_surroundocc_val_rainy} and Table~\ref{tab:nuscenes_surroundocc_val_night}.
MetaOcc-S consistently maintains superior performance across these challenging scenarios, benefiting from the resilience of radar sensing and the effectiveness of the proposed radar feature enhancement and fusion modules.

\begin{figure*}[ht]
    \centering
    \includegraphics[width=\linewidth]{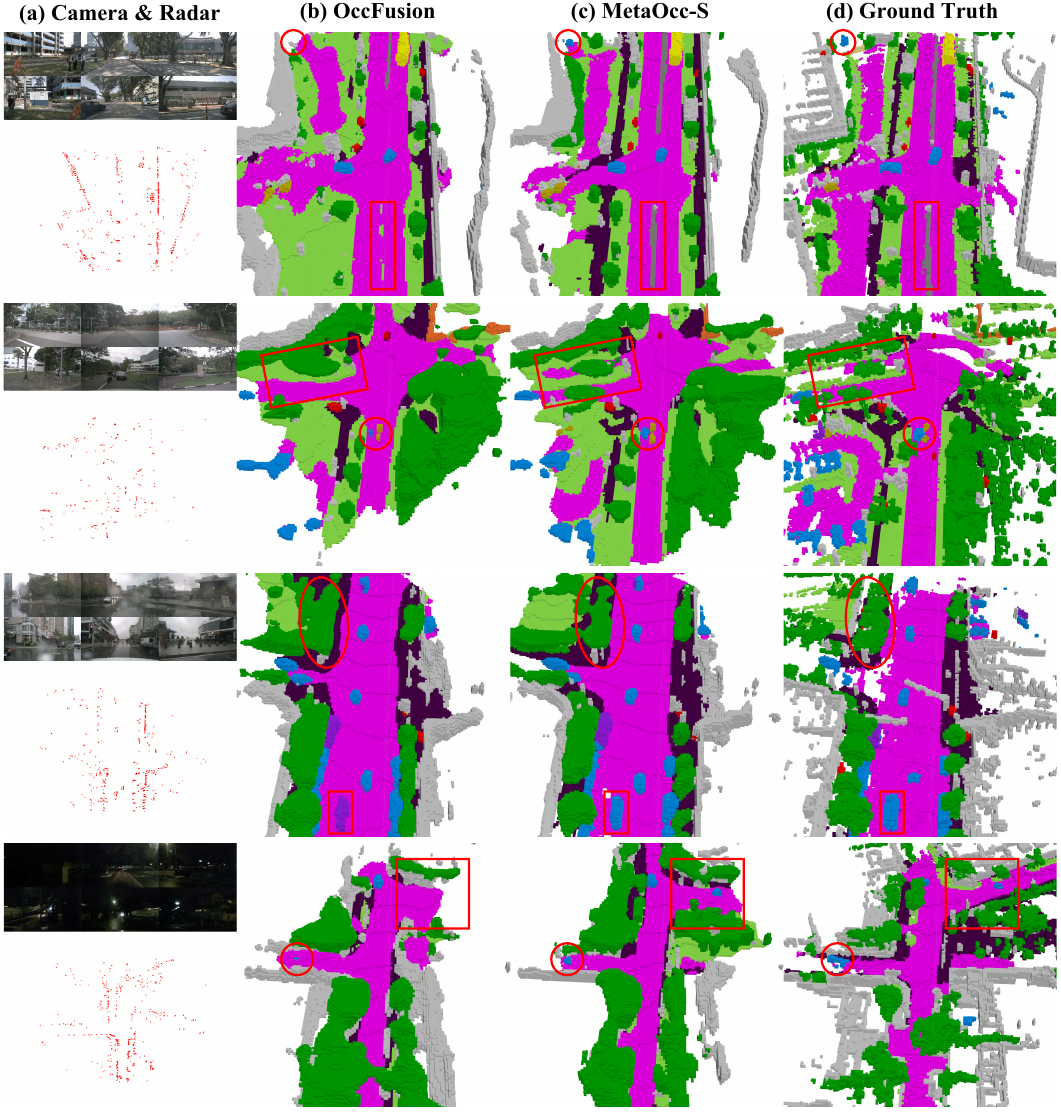}
    \caption{
    \textbf{Qualitative results across diverse scenes on the SurroundOcc-nuScenes dataset}. (a) shows multi-view camera images and radar inputs. (b)–(d) present occupancy predictions from OccFusion, MetaOcc, and the ground truth, respectively. MetaOcc consistently demonstrates enhanced spatial completeness and semantic precision across a variety of real-world driving scenarios.
    }
\label{fig:vis_nuscenes}
\end{figure*}

It is worth noting that the conventional radar used in \textit{SurroundOcc-nuScenes} provides limited vertical resolution, which constrains its utility for detailed 3D occupancy reasoning.
As a result, both MetaOcc-S and OccFusion achieve relatively modest gains compared to camera-only methods in this setting.
Nevertheless, qualitative visualizations in Fig.~\ref{fig:vis_nuscenes} reveal that MetaOcc-S provides enhanced spatial perception and object delineation across diverse environmental conditions, compared to alternative multi-modal fusion baselines.

\definecolor{ncar}{RGB}{255, 165, 0}
\definecolor{npedestrian}{RGB}{128, 0, 128}
\definecolor{nrider}{RGB}{0, 0, 200}
\definecolor{nlarge_vehicle}{RGB}{220, 220, 0}
\definecolor{ncycle}{RGB}{230 ,230 ,230}
\definecolor{nroad_obstacle}{RGB}{255, 69, 0}
\definecolor{ntraffic_fence}{RGB}{0, 0, 150}
\definecolor{ndriveable_surface}{RGB}{135 ,206 ,235}
\definecolor{nsidewalk}{RGB}{200, 200, 200}
\definecolor{nvegetation}{RGB}{34 ,139 ,34}
\definecolor{nmanmade}{RGB}{230, 230, 250}

\setlength{\tabcolsep}{1.4pt}  
\renewcommand{\arraystretch}{1}  

\begin{table*}[htop]
\centering
\caption{
3D semantic occupancy prediction results on the \textbf{OmniHD-Scenes test set}, with models trained on the \textbf{OmniHD-SemiOcc training set} using different proportions of ground-truth (GT) supervision. The GT Ratio denotes the percentage of pseudo-labels substituted with human-annotated ground-truth labels.
}
\label{semi-occ_performance_omnihd_v2}
\resizebox{0.7\textwidth}{!}{
\begin{tabular}{c|c|cc|ccccccccccc}
\hline
\noalign{\smallskip}
GT Ratio               & Method      & SC IoU & mIoU & \rotatebox{90}{\textcolor{ncar}{$\blacksquare$}car}    & \rotatebox{90}{\textcolor{npedestrian}{$\blacksquare$}pedestrian} & \rotatebox{90}{\textcolor{nrider}{$\blacksquare$}rider} & \rotatebox{90}{\textcolor{nlarge_vehicle}{$\blacksquare$}large vehicle} & \rotatebox{90}{\textcolor{ncycle}{$\blacksquare$}cycle} & \rotatebox{90}{\textcolor{nroad_obstacle}{$\blacksquare$}road obstacle} & \rotatebox{90}{\textcolor{ntraffic_fence}{$\blacksquare$}traffic fence} & \rotatebox{90}{\textcolor{ndriveable_surface}{$\blacksquare$}drive. surf.} & \rotatebox{90}{\textcolor{nsidewalk}{$\blacksquare$}sidewalk} & \rotatebox{90}{\textcolor{nvegetation}{$\blacksquare$}vegetation} & \rotatebox{90}{\textcolor{nmanmade}{$\blacksquare$}manmade} \\
\noalign{\smallskip}
\hline
\noalign{\smallskip}
\multirow{3}{*}{0\%}   & SurroundOcc &  28.11 & 10.51 & 18.72 & 4.66 & 9.24  & 15.42 & 0.54 & 0.40 & 2.83  & 40.26 & 6.30  & 14.62 & 2.61 \\
                       & TeOcc       &  32.13 & 12.75 & 23.04 & 3.81 & 16.25 & 18.68 & 1.24 & 0.85 & 5.72  & 41.62 & 4.80  & 18.59 & 5.66 \\
                       & MetaOcc     &  32.38 & 14.46 & 25.59 & 7.92 & 20.92 & 23.22 & 0.96 & 0.77 & 5.28  & 41.10 & 6.49  & 18.71 & 8.09  \\
\noalign{\smallskip}
\hline

\noalign{\smallskip}
\multirow{3}{*}{50\%}  & SurroundOcc & 28.50 & 12.91 & 19.99 & 4.93 & 9.94  & 16.67 & 1.40 & 1.29 & 7.99  & 45.23 & 13.91 & 16.46 & 4.24 \\
                       & TeOcc       &  32.30 & 16.39 & 24.94 & 4.09 & 16.80 & 21.32 & 2.60 & 4.22 & 18.27 & 46.65 & 13.30 & 20.52 & 7.56 \\
                       & MetaOcc     & 32.89 & 19.79 & 26.84 & 7.71 & 21.81 & 24.05 & 2.37 & 2.49 & 32.54 & 49.67 & 16.89 & 21.13 & 12.14 \\
\noalign{\smallskip}
\hline
\noalign{\smallskip}
\multirow{3}{*}{100\%} & SurroundOcc & 28.61 & 15.20 & 21.46 & 3.96 & 10.76 & 16.58 & 1.57 & 2.99 & 21.63 & 48.52 & 18.31 & 16.73 & 4.71 \\
                       & TeOcc       & 32.28 & 17.71 & 26.79 & 5.03 & 19.22 & 21.93 & 1.67 & 2.87 & 26.89 & 47.96 & 13.85 & 21.12 & 7.52 \\
                       & MetaOcc     & 32.75 & 21.73 & 27.52 & 8.84 & 24.06 & 23.35 & 4.79 & 6.47 & 38.75 & 50.29 & 19.91 & 21.88 & 13.22 \\
\noalign{\smallskip}
\hline
\end{tabular}
}
\end{table*}
\setlength{\tabcolsep}{1.4pt}

\definecolor{ncar}{RGB}{255, 165, 0}
\definecolor{npedestrian}{RGB}{128, 0, 128}
\definecolor{nrider}{RGB}{0, 0, 200}
\definecolor{nlarge_vehicle}{RGB}{220, 220, 0}
\definecolor{ncycle}{RGB}{230 ,230 ,230}
\definecolor{nroad_obstacle}{RGB}{255, 69, 0}
\definecolor{ntraffic_fence}{RGB}{0, 0, 150}
\definecolor{ndriveable_surface}{RGB}{135 ,206 ,235}
\definecolor{nsidewalk}{RGB}{200, 200, 200}
\definecolor{nvegetation}{RGB}{34 ,139 ,34}
\definecolor{nmanmade}{RGB}{230, 230, 250}

\setlength{\tabcolsep}{1.4pt}  
\renewcommand{\arraystretch}{1} 

\begin{table*}[htop]
\centering
\caption{
3D semantic occupancy prediction results on the \textbf{OmniHD-Scenes adverse scenario test subset}, with models trained on the \textbf{OmniHD-SemiOcc training set} using varying proportions of ground-truth (GT) supervision. The GT Ratio represents the percentage of pseudo-labels replaced by human-annotated ground-truth labels.
}

\label{semi-occ_performance_omnihd_v2_adverse}
\resizebox{0.7\textwidth}{!}{
\begin{tabular}{c|c|cc|ccccccccccc}
\hline
\noalign{\smallskip}
GT Ratio               & Method      & SC IoU & mIoU & \rotatebox{90}{\textcolor{ncar}{$\blacksquare$}car}    & \rotatebox{90}{\textcolor{npedestrian}{$\blacksquare$}pedestrian} & \rotatebox{90}{\textcolor{nrider}{$\blacksquare$}rider} & \rotatebox{90}{\textcolor{nlarge_vehicle}{$\blacksquare$}large vehicle} & \rotatebox{90}{\textcolor{ncycle}{$\blacksquare$}cycle} & \rotatebox{90}{\textcolor{nroad_obstacle}{$\blacksquare$}road obstacle} & \rotatebox{90}{\textcolor{ntraffic_fence}{$\blacksquare$}traffic fence} & \rotatebox{90}{\textcolor{ndriveable_surface}{$\blacksquare$}drive. surf.} & \rotatebox{90}{\textcolor{nsidewalk}{$\blacksquare$}sidewalk} & \rotatebox{90}{\textcolor{nvegetation}{$\blacksquare$}vegetation} & \rotatebox{90}{\textcolor{nmanmade}{$\blacksquare$}manmade} \\
\noalign{\smallskip}
\hline
\noalign{\smallskip}
\multirow{3}{*}{0\%}   & SurroundOcc &  25.41 & 10.44 & 20.79 & 5.44 & 8.17  & 16.84 & 0.51 & 0.30 & 2.06  & 34.39 & 7.59  & 15.88 & 2.82 \\
                       & TeOcc       &  29.77 & 12.44 & 23.13 & 3.09 & 15.66 & 18.72 & 2.07 & 0.59 & 5.66  & 40.19 & 4.18  & 17.26 & 6.31 \\
                       & MetaOcc     &  30.87 & 14.08 & 25.48 & 7.36 & 18.68 & 24.33 & 1.25 & 0.43 & 4.76  & 38.91 & 6.11  & 18.65 & 8.96  \\
\noalign{\smallskip}
\hline

\noalign{\smallskip}
\multirow{3}{*}{50\%}  & SurroundOcc & 26.38 & 12.77 & 22.10 & 6.30 & 8.06  & 18.61 & 1.18 & 1.23 & 4.74  & 39.50 & 14.72 & 17.78 & 6.27 \\
                       & TeOcc       &  30.03 & 15.59 & 24.83 & 3.23 & 15.54 & 20.73 & 3.89 & 2.38 & 16.90 & 43.52 & 12.69 & 19.58 & 8.23 \\
                       & MetaOcc     & 31.65 & 18.69 & 26.91 & 7.00 & 20.13 & 24.67 & 2.84 & 1.82 & 26.22 & 45.84 & 16.51 & 20.88 & 12.79 \\
\noalign{\smallskip}
\hline
\noalign{\smallskip}
\multirow{3}{*}{100\%} & SurroundOcc & 27.20 & 14.04 & 21.07 & 3.78 & 9.56  & 15.03 & 1.97 & 2.41 & 14.85 & 45.58 & 18.63 & 16.07 & 5.43 \\ 
                       & TeOcc       & 29.96 & 16.30 & 26.34 & 3.97 & 16.96 & 21.82 & 2.58 & 1.47 & 20.37 & 44.23 & 13.44 & 20.01 & 8.07 \\
                       & MetaOcc     & 31.43 & 20.62 & 27.64 & 8.24 & 22.03 & 23.94 & 6.21 & 4.54 & 31.72 & 46.74 & 20.34 & 21.56 & 13.85 \\
\noalign{\smallskip}
\hline
\end{tabular}
}
\end{table*}
\setlength{\tabcolsep}{1.4pt}

\begin{figure}[ht]
    \centering
    \includegraphics[width=0.48\textwidth]{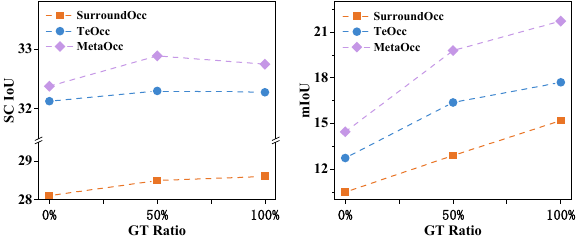}
    \caption{
    SC IoU and mIoU comparison of different models trained with varying proportions of ground-truth (GT) labels on the \textbf{OmniHD-Scenes test set}.
    }
    \label{fig:vis_performance_semiocc}
\end{figure}

\begin{figure}[h]
    \centering
\includegraphics[width=0.48\textwidth]{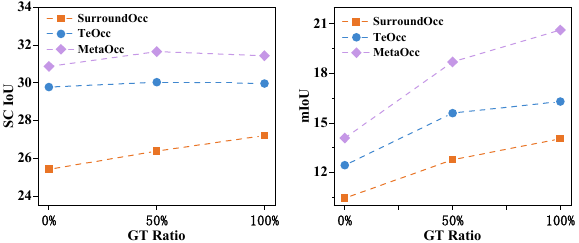}
    \caption{
    SC IoU and mIoU comparison of different models trained with varying proportions of ground-truth (GT) labels on the \textbf{OmniHD-Scenes adverse scenario subset}.
    }
    \label{fig:vis_performance_semiocc_adverse}
\end{figure}

\textbf{Semi-Supervised Results on OmniHD-Scenes.}
The results of semi-supervised occupancy prediction across different scenarios under varying ground-truth (GT) supervision levels are presented in Table~\ref{semi-occ_performance_omnihd_v2} and Table~\ref{semi-occ_performance_omnihd_v2_adverse}.
Corresponding qualitative results are shown in Fig.\ref{fig:vis_performance_semiocc} and Fig.\ref{fig:vis_performance_semiocc_adverse}, which visualize the impact of different GT-to-pseudo-label ratios on model performance.

Both GT and pseudo-labels are generated using the same LiDAR point clouds for geometric reconstruction, ensuring consistent spatial alignment across annotations.
As a result, models exhibit stable geometric performance in terms of SC IoU across all settings, as reflected in the visualizations.
The key distinction between the two label types lies in the source of semantic information: pseudo-labels are derived from automated vision-language segmentation models, while GT labels are manually annotated for static LiDAR points.
This difference significantly influences semantic prediction quality, particularly for fine-grained categories.

As reported in Table~\ref{semi-occ_performance_omnihd_v2}, when training with only pseudo-labels (GT Ratio = 0\%), the mIoU of the camera-only model SurroundOcc reaches 10.51, while MetaOcc achieves 14.46, approximately 65\% of their fully supervised performance.
This demonstrates the effectiveness of the proposed pseudo-label generation pipeline.
Furthermore, when the GT ratio increases to 50\%, mIoU improves by approximately 2.4 for SurroundOcc and 5.33 for MetaOcc, bringing both models to roughly 90\% of their fully supervised levels. These findings collectively indicate that a modest proportion of ground-truth supervision substantially enhances feature learning and prediction accuracy.

As shown in Table~\ref{semi-occ_performance_omnihd_v2_adverse}, both models retain strong performance even under adverse environmental conditions, validating the robustness of the semi-supervised framework.

In summary, the proposed pseudo-labeling strategy significantly reduces the cost of semantic point cloud annotation while preserving a high level of predictive performance.
Moreover, introducing a moderate amount of ground-truth supervision yields notable semantic improvements, offering a practical and cost-effective training paradigm for real-world 3D scene understanding.

\subsection{Ablation Study}
We conduct ablation studies on the \textit{OmniHD-Scenes} test set to assess the contribution of each individual component in the MetaOcc framework.
The baseline configuration employs a ResNet-50 backbone, an input image resolution of $544 \times 960$, and three-frame temporal fusion.

\setlength{\tabcolsep}{3pt}
\renewcommand{\arraystretch}{1}

\begin{table}[t]
    \centering
    \caption{Ablation study of the main components of MetaOcc.}
    \begin{tabular}{c c c|c c}
        \toprule
        \makebox[1.0cm]{RHS} & \makebox[1.0cm]{LAF} & \makebox[1.0cm]{GCF} & SC IoU & mIoU \\
        \midrule
        & & & 31.12 & 19.60 \\
        \text{\checkmark} & & & 31.93 & 20.95 \\
        \text{\checkmark} & \text{\checkmark} & & 32.48 & 21.36 \\
        \text{\checkmark} & \text{\checkmark} & \text{\checkmark} & 32.75 & 21.73 \\
        \bottomrule
    \label{tab:ablation_MetaOcc}
    \end{tabular}
\end{table}

\textbf{Main Components.}
MetaOcc is constructed by progressively integrating three key modules into the radar-camera fusion baseline: Radar Height Self-Attention (RHS), Local Adaptive Fusion (LAF), and Global Cross-Attention Fusion (GCF).

As shown in Table~\ref{tab:ablation_MetaOcc}, the baseline model without any proposed enhancements achieves 31.12 SC IoU and 19.60 mIoU.
Incorporating the RHS module results in a notable improvement of +0.81 SC IoU and +1.35 mIoU, validating the effectiveness of vertical spatial reasoning in enhancing radar feature representation.
Adding the LAF module further improves performance by +0.55 SC IoU and +0.41 mIoU, demonstrating the benefits of adaptive voxel-wise cross-modal fusion.
Finally, introducing the GCF module, which enables global radar-camera interaction via deformable attention, yields an overall gain of +1.63 SC IoU and +2.13 mIoU relative to the baseline.

In summary, each proposed module contributes positively and consistently to the overall performance.
Their combined integration results in substantial improvements in both semantic segmentation and geometric occupancy prediction, thereby validating the architectural design of MetaOcc.

\setlength{\tabcolsep}{3pt}
\renewcommand{\arraystretch}{1}

\begin{table}[t]
    \centering
    \caption{Ablation study of Radar Height Self-Attention module.}
    \begin{tabular}{l|cc} 
        \toprule
        Method & SC IoU & mIoU \\
        \midrule
        Second \cite{yan2018second} & 32.52 & 21.34 \\
        PointPillars + Expand & 32.17 & 21.64 \\
        PointPillars + RHS (ours) & 32.75 & 21.73 \\
        \bottomrule
    \end{tabular}
    \label{tab:ablation_RHS}
\end{table}

\textbf{Radar Height Self-Attention.}
To investigate radar feature extraction strategies, we compare three model variants:
(1) a baseline using the commonly adopted SECOND encoder \cite{yan2018second}, originally designed for LiDAR, which extracts voxel-based 3D features;
(2) PointPillars with expansion operators for naive height reconstruction; and
(3) our proposed PointPillars combined with RHS, which explicitly models vertical feature distributions.

As reported in Table~\ref{tab:ablation_RHS}, the RHS-enhanced model achieves the best overall performance.
Compared to the SECOND-based baseline, RHS improves mIoU by +0.39, indicating enhanced semantic feature extraction.
Relative to PointPillars with expand operators, RHS increases SC IoU by +0.58, confirming its effectiveness in geometric reasoning through height-aware spatial modeling.

These results demonstrate that RHS successfully combines the geometric structure modeling capabilities of voxel-based encoders with the semantic representation strength of pillar-based methods.
Its targeted design for sparse radar point clouds makes it more suitable than conventional LiDAR-oriented backbones for 3D occupancy prediction tasks.

\setlength{\tabcolsep}{3pt}
\renewcommand{\arraystretch}{1}

\begin{table}[t]
    \centering
    \caption{
    Ablation study of the Global Cross-Attention Fusion module.
    }
    \begin{tabular}{c|cc} 
        \toprule
        \makebox[3cm]{Method} & SC IoU & mIoU \\
        \midrule
        Default Initialization & \underline{32.28} & 21.34 \\
        LAF Initialization & 32.75 & \underline{21.73} \\
        \bottomrule
    \end{tabular}
    \label{tab:ablation_GCF}
\end{table}

\textbf{Global Cross-Attention Fusion.}
We examine the effect of query feature initialization in the GCF module.
The baseline configuration uses randomly initialized queries, whereas our method utilizes fused multi-modal features from the LAF module to guide the cross-attention mechanism.

As shown in Table~\ref{tab:ablation_GCF}, initializing GCF with LAF-derived features yields performance gains of +0.39 mIoU and +0.47 SC IoU compared to random initialization.
These improvements suggest that leveraging prior fusion knowledge helps enhance both fusion quality and cross-modal alignment during global attention.

Overall, the results confirm the effectiveness of LAF-informed query initialization in boosting the capacity of the GCF module for robust and context-aware feature interaction.

\setlength{\tabcolsep}{4pt}
\renewcommand{\arraystretch}{1}

\begin{table}[t]
    \centering
    \caption{
    Ablation study of the Temporal Fusion module.
    }
    \begin{tabular}{l|cc|cc}
        \toprule
        Length & SC IoU & mIoU & GPU Mem. & FPS \\
        \midrule
        T = 1 & 31.52 & 20.92 & 15.1 GB & 3.5 \\ 
        T = 2 & 32.48 & 21.38 & 16.3 GB & 3.4\\
        T = 3 & 32.75 & 21.73 & 18.5 GB & 3.3 \\
        T = 4 & 32.83 & 22.40 & 19.9 GB & 3.1 \\
        \bottomrule
    \end{tabular}
    \label{tab:ablation_TAF}
\end{table}

\textbf{Temporal Fusion.}
To evaluate the impact of temporal fusion, we vary the number of temporal multi-modal feature frames from 1 to 4, while maintaining the same model architecture.

As presented in Table~\ref{tab:ablation_TAF}, increasing the number of frames consistently improves performance, with SC IoU and mIoU rising by 1.31 and 1.48, respectively.
These results confirm that incorporating temporal information enhances the model’s ability to capture dynamic scene context and improves both geometric and semantic understanding.
However, the benefits come at the cost of increased computational demands: training-time GPU memory usage rises, reaching 19.9 GB at 4 frames, and the inference speed decreases to 3.1 FPS.
This highlights the trade-off between temporal modeling accuracy and computational efficiency in real-world deployment scenarios.

\setlength{\tabcolsep}{3pt}
\renewcommand{\arraystretch}{1}

\begin{table}[t]
    \centering
    \caption{
    Ablation study on image resolution.
    }
    \begin{tabular}{c|cc} 
        \toprule
        \makebox[3cm]{Image Resolution} & SC IoU & mIoU \\
        \midrule
        352×576 & 32.63 & 21.01 \\
        544×960 & 32.75 & 21.73 \\
        864×1536 & 33.22 & 22.34 \\
        \bottomrule
    \end{tabular}
    \label{tab:ablation_img_res}
\end{table}

\textbf{Image Resolution.}
We further evaluate the impact of input image resolution on 3D occupancy prediction performance.
Table~\ref{tab:ablation_img_res} compares three configurations: 352×576, 544×960, and 864×1536.
As the resolution increases, both SC IoU and mIoU improve consistently, with the highest resolution achieving 33.22 SC IoU and 22.34 mIoU.
These results indicate that higher-resolution images provide richer visual cues and finer-grained semantic details, which contribute to more accurate spatial reasoning and semantic understanding in occupancy prediction.

\section{Conclusions}
\label{sec:conclusions}
This paper introduces MetaOcc, a novel multi-modal 3D occupancy prediction framework that integrates surround-view cameras and 4D radar to enhance scene perception for autonomous driving. To address the challenges of sparse and noisy radar signals, we propose a Radar Height Self-Attention module that explicitly models vertical spatial distributions, enabling improved geometric reasoning from radar point clouds. In parallel, the Hierarchical Multi-scale Multi-modal Fusion strategy facilitates both local and global alignment across modalities and time, leading to more comprehensive and robust occupancy representations.
Extensive experiments on the OmniHD-Scenes and SurroundOcc-nuScenes benchmarks demonstrate that MetaOcc achieves state-of-the-art results in both semantic and geometric metrics. In particular, the model exhibits notable improvements in detecting small and partially occluded objects and retains strong performance under challenging conditions such as rain and nighttime driving. These findings underscore the benefits of radar-visual complementarity and the effectiveness of the proposed fusion mechanisms.

Beyond fully supervised settings, we introduce a cost-efficient pseudo-labeling pipeline and a semi-supervised training strategy that significantly reduce the dependency on expensive 3D point cloud annotations. Our results show that even with limited or no ground-truth labels, the proposed method maintains high accuracy, confirming the feasibility of scaling 3D occupancy prediction systems with minimal human labeling effort.

Collectively, these contributions demonstrate that MetaOcc not only advances the accuracy and robustness of multi-modal occupancy perception but also improves the practicality and scalability of such systems in real-world deployment. Future work may explore the integration of additional modalities and improved uncertainty modeling to further extend the applicability of the proposed framework.

\bibliographystyle{elsarticle-num-names}
\bibliography{IEEEexample}

\appendix

\end{document}